%% file: main.tex
\documentclass{article}

\usepackage{microtype}
\usepackage{graphicx}
\usepackage{booktabs} 
\usepackage{caption}
\usepackage{subcaption}
\usepackage{hyperref}

\usepackage[accepted]{icml2025}

\usepackage{amsmath}
\usepackage{amssymb}
\usepackage{mathtools}
\usepackage{amsthm}

\usepackage{url}
\usepackage{algorithm}
\usepackage{algorithmic}
\usepackage{multirow}
\usepackage{array}
\input{math_commands.tex}

\usepackage[capitalize,noabbrev]{cleveref}

\theoremstyle{plain}
\newtheorem{theorem}{Theorem}[section]
\newtheorem{proposition}[theorem]{Proposition}

\theoremstyle{definition}

\theoremstyle{remark}

\usepackage[textsize=tiny]{todonotes}
\usepackage[normalem]{ulem}

\newcommand{\mean}[2]{\mathbb{E}_{#2} \! \left[ #1 \right]}
\newcommand{\norm}[1]{\left\lVert#1\right\rVert}
\newcommand{\vect}[1]{\boldsymbol{#1}}

\usepackage{colortbl}  

\icmltitlerunning{VCT: Training Consistency Models with Variational Noise Coupling}

\begin{document}

\twocolumn[
\icmltitle{VCT: Training Consistency Models with Variational Noise Coupling}



\icmlsetsymbol{equal}{*}

\begin{icmlauthorlist}
\icmlauthor{Gianluigi Silvestri$^*$}{oneplanet,donders}
\icmlauthor{Luca Ambrogioni}{donders}
\icmlauthor{Chieh-Hsin Lai}{sony}
\icmlauthor{Yuhta Takida}{sony}
\icmlauthor{Yuki Mitsufuji}{sony,sony2}
\end{icmlauthorlist}

\icmlaffiliation{oneplanet}{OnePlanet Research Center, imec-the Netherlands, Wageningen, the Netherlands}
\icmlaffiliation{donders}{Donders Institute for Brain, Cognition and Behaviou, Nijmegen, the Netherlands}
\icmlaffiliation{sony}{Sony AI, Tokyo, Japan}
\icmlaffiliation{sony2}{Sony Group Corporation}

\icmlcorrespondingauthor{Gianluigi Silvestri}{gianluigi.silvestri@imec.nl, gianlu.silvestri@gmail.com}

\icmlkeywords{Machine Learning, ICML, Generative Models, Consistency Models, Variational Inference}

\vskip 0.3in
]



\printAffiliationsAndNotice{$^*$Work done during an internship at Sony AI} 

\begin{abstract}


Consistency Training (CT) has recently emerged as a strong alternative to diffusion models for image generation. However, non-distillation CT often suffers from high variance and instability, motivating ongoing research into its training dynamics. We propose Variational Consistency Training (VCT), a flexible and effective framework compatible with various forward kernels, including those in flow matching. Its key innovation is a learned noise-data coupling scheme inspired by Variational Autoencoders, where a data-dependent encoder models noise emission. This enables VCT to adaptively learn noise-to-data pairings, reducing training variance relative to the fixed, unsorted pairings in classical CT.
 Experiments on multiple image datasets demonstrate significant improvements: our method surpasses baselines, achieves state-of-the-art FID among non-distillation CT approaches on CIFAR-10, and matches SoTA performance on ImageNet $64 \times 64$ with only two sampling steps. Code is available at \url{https://github.com/sony/vct}.

\end{abstract}

\begin{figure}[ht!]
    \centering
    \includegraphics[width=\linewidth]{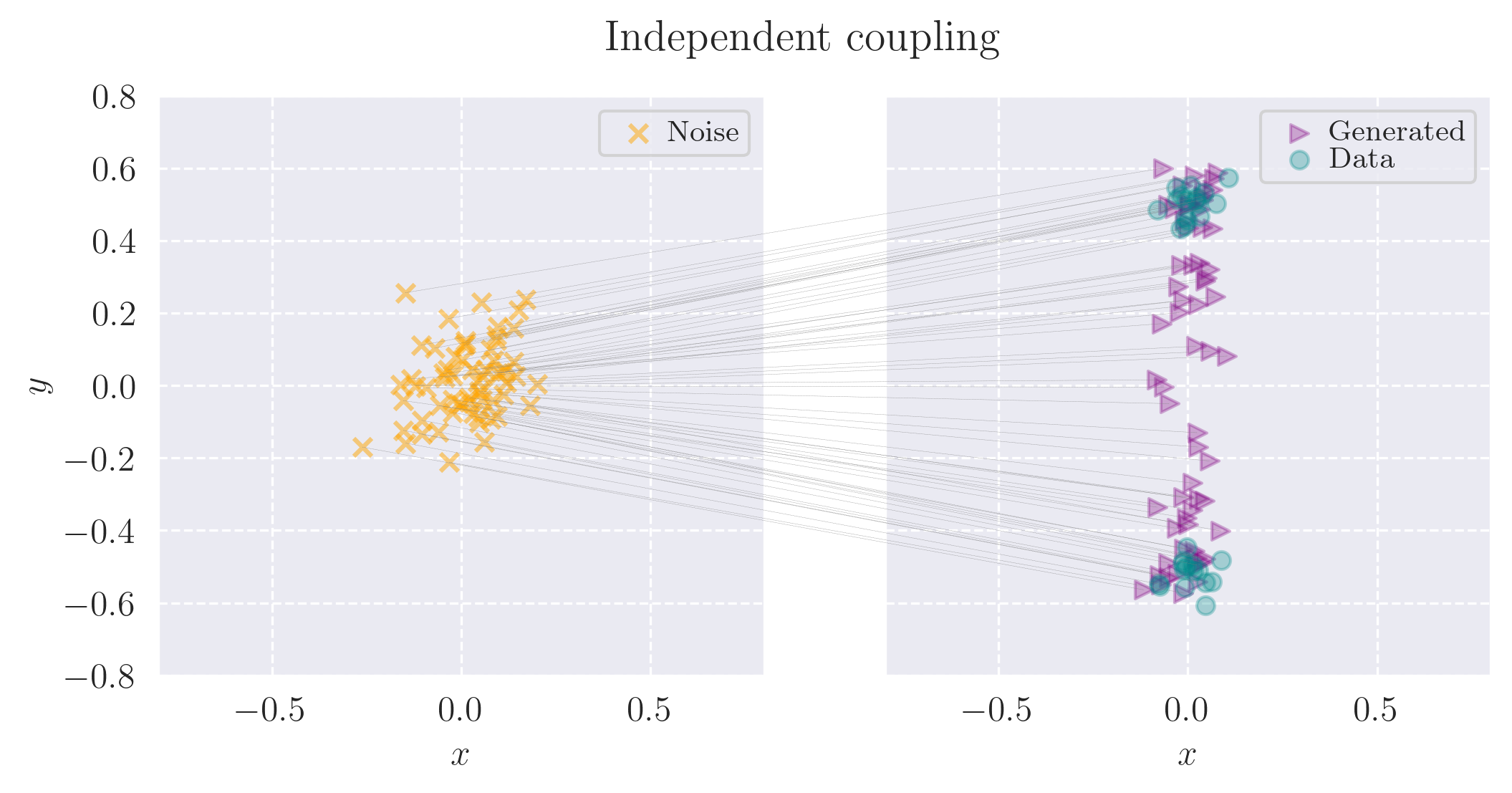}
    \includegraphics[width=\linewidth]{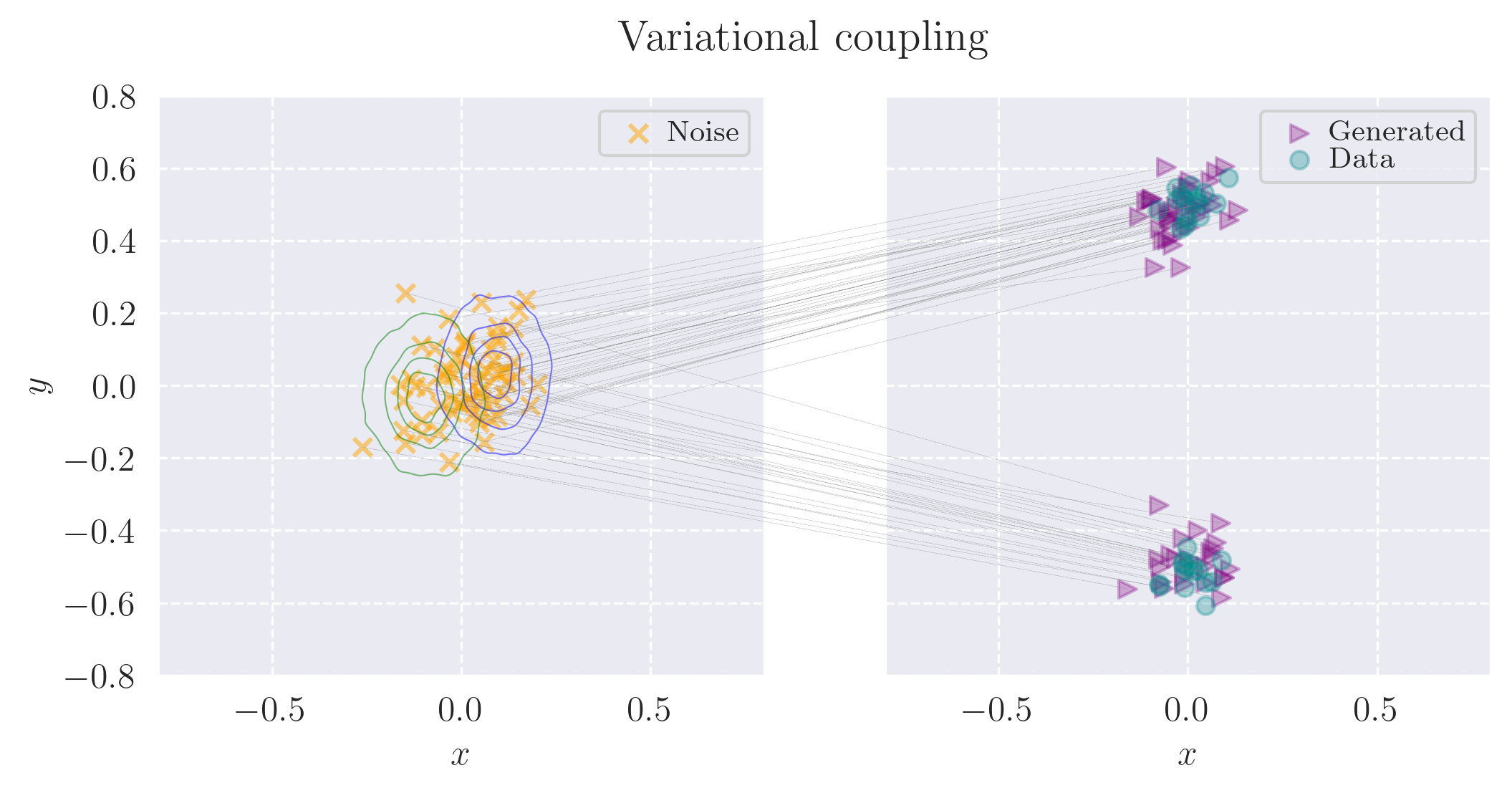}
    \caption{Comparison of 1-step generation on toy data for independent and variational coupling. The data is sampled from a 2-d mixture of Gaussians with means $\vect{\mu}_1=(0 , 0.5)$ and $\vect{\mu}_2=(0, -0.5)$. On the bottom plot (our approach), we show the posterior probabilities learned by the encoder (in blue and green) corresponding to $p(\vect{z}\mid \vect{\mu}_1)$ and $p(\vect{z}\mid \vect{\mu}_2)$, and their cumulative sum approximately recovers the prior distribution. The gray lines connect the samples from the trained models to the corresponding input noise. More details about the toy experiments are given in Appendix \ref{app:toy}.}
    \label{fig:toy_samples_1}
\end{figure}
\section{Introduction}
Generative Models are deep learning algorithms designed to learn the underlying probability distribution of a given dataset, in order to then generate samples coming from such a distribution. Some widely used models are Generative Adversarial Networks \citep{goodfellow2014generative}, Variational Autoencoders \citep[VAE;][]{kingma2013auto, rezende2014stochastic}, and Normalizing Flows \citep{chen2018neural, papamakarios2021normalizing, kobyzev2020normalizing}. More recently, Diffusion Models \citep{sohl2015deep, ho2020denoising, songscore} have achieved state-of-the-art (SoTA) results in several domains, including images, videos, and audio \citep{dhariwal2021diffusion, rombach2022high, karras2022elucidating, karras2024analyzing, ho2022video, kongdiffwave}. However, a weakness of diffusion models is the need for an iterative sampling procedure, which can require hundreds of network evaluations. Therefore, substantial effort has been made to develop methods that can maintain similar generation quality while requiring fewer sampling iterations \citep{songdenoising, jolicoeur2021gotta, salimansprogressive, liupseudo, lu2022dpm}. Among such methods, a recent and promising direction is given by Consistency Models (CMs) \citep{song2023consistency}. CMs, while sharing many similarities with DMs, use a different training procedure as they directly learn the probability flow equations rather than the score function. CMs can be either trained by distilling the ODE trajectories of a pre-trained diffusion model (Consistency Distillation, CD), or completely from scratch through a bootstrap loss (Consistency Training, CT), which results in a novel generative modeling framework. However, the CT objective can be subject to high variance, making it difficult to train. A follow-up work \citep{songimproved} analyses the training dynamics of CMs and proposes several improvements which result in a more stable CT procedure, achieving SoTA results in few-step image generation. Since then, several works have proposed additional strategies to further improve CT \citep{geng2024consistency, wang2024stable, lee2024truncated, yang2024consistency}. 


A possible source of instability of CT training comes from the fact that different noise masks are applied to the same data point, creating ambiguity the target corresponding to the given noisy state, especially early during training with coarse discretization steps. From a more mathematical perspective, training can be destabilized by sharp boundaries in the ODE flow mapping, which can be hard to learn for the model and give raise to high variance in the stochastic gradient estimator. For example, the ODE flow for mixture of delta distributions is defined by a tessellation of the initial noise space, with discontinuities along all the borders. Since the standard CT approach with fixed forward process cannot alter the target ODE flow, the optimum of the standard CT training can potentially be highly singular. The existence of these singularities depend on topological reasons (i.e. non-injectivity of the ODE flow mapping at $t \rightarrow 0$ \citep{cornish2020relaxing}). However, the issue can likely be ameliorated by altering the forward process during training, which can be used to change the location of the singularities.

A way to implement this approach is to sample noise using a conditional coupling function between data and noise. The concept of using a coupling function to reduce variance during training was successfully used in Flow Matching, with works such as \citep{pooladian2023multisample, tong2023conditional, lee2023minimizing}, with the main objective of obtaining straighter ODE trajectories for faster sampling. Forms of coupling in CMs were proposed in works such as \citep{douunified, issenhuth2024improving}, but their formulations do not match the performance of standard CMs. 

In this work, we propose \emph{Variational Consistency Training (VCT)}, which introduces a variational formulation for training the forward transition kernel via a coupling function between data and noise, a technique we term \emph{Variational Coupling (VC)}. This yields a loss function analogous to that of VAEs. By learning a data-dependent noise distribution regularized with an additional Kullback-Leibler (KL) divergence term, VCT enables an end-to-end training procedure compatible with various forward kernels.
 We demonstrate that this learned coupling effectively improves generation performance and scales well to high-dimensional data.
 A simple and intuitive example is shown in Figure \ref{fig:toy_samples_1}, where with the same settings, the model trained with the learned coupling generates samples closer to the data distribution. From this figure, we can see how the learned coupling partitions the data differently from how the standard CT does, effectively changing the form of the underlying ODE, likely resulting in an easier training objective as the prior noise is partitioned according to the learned the data-dependent coupling distribution. 

The reminder of the paper is organized as follows: we first discuss relevant related CT methods and flow-based works employing coupling strategies. We then formulate CT from the Flow Matching perspective, which is a generalization of the diffusion framework and it offers a more natural way to introduce the noise-coupling distribution. Finally, we describe our method, deriving similarities with Variational Autoencoders, and report our experimental results on common image benchmarks.



\section{Related Work}
Since the introduction of Consistency Models in \citep{song2023consistency, songimproved}, several strategies have been proposed to improve training stability. The work from \citep{geng2024consistency} proposes Easy Consistency Models (ECM) a novel training strategy where time steps are sampled in a continuous fashion and the discretization step is adjusted during training, as opposed to the discrete time grid used in iCT. It further shows the benefits of initializing the network weights with the ones from a pretrained score model, achieving superior performance with smaller training budget. \citep{wang2024stable} builds on top of ECM, introducing additional improvements and framing consistency training as value estimation in Temporal Difference learning \citep{sutton2018reinforcement}. Truncated consistency models, introduced in \citep{lee2024truncated}, proposes to add a second training stage on top of ECM, to allow the model to focus its capacity on the later time steps, resulting in improved few-steps generation performance. Other recent contributions to the consistency model literature are works such as \citep{kimconsistency, heek2024multistep} where the focus is on improving multistep sample quality, \citep{lee2024stabilizing} which trains a model with both consistency and score loss to reduce variance, and \citep{lu2024simplifying}, which proposes several improvements to the continuous-time training of consistency models. Our work can be seen as a parallel contribution to the aforementioned methods, as we focus on learning the data-noise coupling, which can be used as drop-in replacement to the standard independent coupling.

There are several works showing the benefit of using coupling in Flow Matching \citep{pooladian2023multisample, tong2023conditional, liu2023flow, lee2023minimizing, albergostochastic, kim2024simple} and Diffusion Models \citep{bartosh2024neural1, bartosh2024neural2, nielsen2024diffenc}. Among these, \citep{lee2023minimizing} shares the most similarities with our method, as they also use an encoder to learn a probability distribution over the noise conditioned on the data. Their method results in improved performance compared to equivalent Flow Matching models, while requiring less function evaluations. Our method consists of a similar procedure but applied to CT, resulting in improved few-steps generation performance and confirming the effectiveness of learning the data-noise coupling. A different coupling strategy for CT is proposed in \citep{issenhuth2024improving}, where the data-noise coupling is extracted directly from the prediction of the consistency model during training. Compared to our method, they do not need the additional encoder to learn the coupling, but their generator-induced coupling needs to be alternated with the standard independent coupling to avoid instabilities. The Flow Matching formulation in Consistency Models with linear interpolation kernel was previously used in \citep{douunified, yang2024consistency}, where the former also explores the use of minibatch OT coupling, while the latter trains the model to learn the velocity field and adds a regularization term to enforce constant velocity. In our work, we use the Flow Matching formulation, but keep most of the CT building blocks, resulting in a simpler formulation with superior performance.

\section{Background} 

Flow Matching provides a general framework that generalizes diffusion and score-matching models \citep{lipmanflow, albergo2023building}. In the Flow Matching formalism, a deterministic flow function $\vect{\psi}_t$ with initial condition $\rvx_0$ is used to build an interpolating map between two distributions $\vect{\psi}_t(\vect{x}_0) = \vect{x}_t$, such that the data distribution $p_0(\vect{x}_0)$ is mapped into a distribution $p_1(\vect{x}_1)$, commonly chosen to be Gaussian noise distribution $p_1(\vect{x}_1):=\mathcal{N}(\vect{x}_1;\bm{0},\vect{I})$,  by the pushforward operator. From this quantity, we can define the vector field $\vect{u}_t(\vect{x}_t)$ as the infinitesimal generator of $\vect{\psi}_t$:
\begin{align*}
    \frac{\mathrm{d}}{\mathrm{d} t} \vect{\psi}_t(\vect{x}_0) &= \vect{u}_t(\vect{x}_t) = \vect{u}_t(\vect{\psi}_t(\vect{x}_0)) ~, 
\end{align*}
In a diffusion model, the flow $\vect{\psi}_t(\vect{x}_0)$ is the inverse of the probability ODE flow determined by the forward SDE. Instead, in standard Flow Matching, the mapping is specified as a conditional flow $\vect{\psi}_t(\vect{x}_0; \vect{x}_1)$, which is typically taken as a simple linear interpolation between samples from the two densities $p_0(\vect{x}_0)$ and $p_1(\vect{x}_1)$. Some common examples are $\vect{\psi}_t(\vect{x}_0; \vect{x}_1) = \vect{x}_t = (1-t) \vect{x}_0 + t\vect{x}_1$ as seen in \citep{lipmanflow}, and $\vect{\psi}_t(\vect{x}_0; \vect{x}_1) = \vect{x}_t = \vect{x}_0 + t\vect{x}_1$ as in \citep{karras2022elucidating}. This conditional flow is analogous to the formal solution kernel of the forward process at time $t$ in the generative diffusion framework. While it is difficult to directly obtain the flow function $\vect{\psi}_t(\vect{x}_0)$ from the conditional flow $\vect{\psi}_t(\vect{x}_0; \vect{x}_1)$, it is possible to give a formal expression for the resulting vector field:
\begin{align}\label{eq: velocity from conditional}
    \vect{u}_t(\vect{x}_t) =  \mean{\vect{u}_t(\vect{x}_t; \vect{x}_1)}{\vect{x}_1 \mid \vect{x_t}}~,
\end{align}
where $\vect{u}_t(\vect{x}_t; \vect{x}_1) = \frac{\mathrm{d}}{\mathrm{d} t} \vect{\psi}_t(\vect{x}_0; \vect{x}_1)$ is the vector field that generates the conditional flow $\vect{\psi}_t(\vect{x}_0; \vect{x}_1)$. In the case of the simple interpolation conditional flows $\vect{\psi}_t(\vect{x}_0; \vect{x}_1) = (1-t) \vect{x}_0 + t \vect{x}_1$, the formula specializes as follows:
\begin{align}
    \vect{u}_t(\vect{x}_t) =  \mean{\vect{x}_1 - \vect{x}_0}{\vect{x}_1 \mid \vect{x_t}} \,.
\end{align}
Readers who are familiar with generative diffusion will immediately recognize that this expression is directly related to the standard expression for the score function. From this connection, it is clear that the conditional vector field can be estimated with a regression objective
\begin{align*}
    \mathbb{E}_{t,\vect{x}_0, \vect{x}_1} || \vect{f}_{\theta}(\vect{\psi}_t(\vect{x}_0; \vect{x}_1), t) - \vect{u}_t(\vect{\psi}_t(\vect{x}_0; \vect{x}_1) ; \vect{x}_1)||^2_2.
\end{align*}

\subsection{Noise coupling}
An advantage of the Flow Matching formalism over SDE diffusion is that, as shown in \citep{pooladian2023multisample, tong2023conditional}, it is straightforward to introduce a probabilistic coupling $
\pi(\vect{x}_1 \mid \vect{x}_0)$ between the data and the noise distribution. In this case, we require that $\int \pi(\vect{x}_1 \mid \vect{x}_0) \mathrm{d} \vect{x}_0$ should follow a standard normal distribution, at least approximately. 
The use of a non-trivial noise coupling does not alter the form of the conditional velocity fields $\vect{u}_t(\vect{x}_t; \vect{x}_1)$ as far as the coupling is time-independent. However, it does alter the total velocity field $\vect{u}_t(\vect{x}_t)$ since it affects the conditional distribution $p(\vect{x}_1, \vect{x}_t)$~, which determines the expectation in Eq.~\ref{eq: velocity from conditional}.


\section{Continuous consistency models from a Flow Matching perspective}
As explained above, the flow function $\vect{\psi}_t(\vect{x}_0)$ maps a noiseless state $\vect{x}_0$ to the noisy state $\vect{x}_t$. Its inverse $\vect{\psi}_t^{-1}(\vect{x}_t)$ can then be interpreted as a denoiser, as it maps each noisy state to a uniquely defined noiseless state $\vect{x}_0$. This function is often referred to as a consistency map, and it follows the identity:
\begin{align} \label{eq: total derivative}
    \frac{\mathrm{d}}{\mathrm{d}t} \vect{\psi}_t^{-1}(\vect{\psi}_t(\vect{x}_0))=0, \quad \text{on}\quad t\in[0,1],
\end{align}
together with the boundary condition $\vect{\psi}_0^{-1}(\vect{x}_0) = \vect{x}_0$. Eq.~\ref{eq: total derivative} is a consequence of the fact that all noisy states in an ODE trajectory $\vect{\psi}_t(\vect{x}_0)$ share the same initial point $\vect{x}_0$, which implies that $\vect{\psi}_t^{-1}(\vect{\psi}_t(\vect{x}_0))$ is constant along the trajectory. This property can be used to define a continuous loss for a network $\vect{f}_{\vect{\theta}}(\vect{x}_t, t)$, trained to approximate the  inverse flow $\vect{\psi}_0^{-1}(\vect{x}_t)$:

\begin{align} \label{eq: continuous loss}
    \mathcal{L}_{\text{cont}}^{\text{tot}}(\vect{\theta}) \equiv  \mean{\int_0^1 \lambda(t) \norm{\frac{\mathrm{d}}{\mathrm{d}t} \vect{f}_{\vect{\theta}}(\vect{\psi}_t(\vect{x}_0), t)}^2 \mathrm{d} t}{\vect{x}_0} 
\end{align}
where $\lambda(t)$ is a positive-valued function that weights the loss for different time points. This loss should be used together with the identity boundary condition $\vect{f}_{\vect{\theta}}(\vect{x}_0, 0) = \vect{x}_0$, which we will discuss later.
 In distillation training, the deterministic trajectories $\vect{x}_t = \vect{\psi}_t(\vect{x}_0)$ are obtained by integrating the ODE flow obtained from a pre-trained diffusion or flow matching model. Alternatively, the consistency network can be trained directly by re-writing the total derivative in terms of the conditional flow: 
\begin{align} \label{eq: conditional consistency}
\begin{split}
    & \frac{\mathrm{d}}{\mathrm{d}t} \vect{\psi}_t^{-1}(\vect{\psi}_t(\vect{x}_0)) = \nabla \vect{\psi}_t^{-1}(\vect{x}_t)   \frac{\mathrm{d} x_t}{\mathrm{d} t} +  \partial_t \vect{\psi}_t^{-1}(\vect{x}_t) \\
    & = \nabla \vect{\psi}_t^{-1}(\vect{x}_t)   \mean{\vect{u}_t(\vect{x}_t; \vect{x}_1)}{\vect{x}_1 \mid \vect{x_t}} +  \partial_t \vect{\psi}_t^{-1}(\vect{x}_t) \\
    & = \mean{\nabla \vect{\psi}_t^{-1}(\vect{x}_t)   \vect{u}_t(\vect{x}_t; \vect{x}_1) +  \partial_t \vect{\psi}_t^{-1}(\vect{x}_t)}{\vect{x}_1 \mid \vect{x_t}} \\
    & = \mean{\frac{\mathrm{d}}{\mathrm{d}t} \vect{\psi}_t^{-1}(\vect{\psi}_t(\vect{x}_0; \vect{x}_1))}{\vect{x}_1 \mid \vect{x_t}}~.
\end{split}
\end{align}
From this equality, together with the fact that the squared Euclidean norm is a convex function, it follows that
\begin{align*}
    \begin{split}
     &\mathcal{L}_{\text{cont}}^{\text{tot}}(\vect{\theta}) \leq \mathcal{L}_{\text{cont}}^{\text{cond}}(\vect{\theta})~, ~~~~\text{with}\\ 
     & \mathcal{L}_{\text{cont}}^{\text{cond}}(\vect{\theta}) \equiv \mean{ \int_0^1 \lambda(t) \norm{\frac{\text{d}}{\text{d}t} \vect{f}_{\vect{\theta}}(\vect{\psi}_t(\vect{x}_0; \vect{x}_1), t)}^2 \mathrm{d} t}{\vect{x}_1, \vect{x_0}}~,
     \end{split}
\end{align*}
where we moved the expectation outside of the squared norm using Jensen's inequality. Therefore, we can optimize the tractable "conditional loss" $\mathcal{L}_{\text{cont}}^{\text{cond}}(\vect{\theta})$ instead of $\mathcal{L}_{\text{cont}}^{\text{tot}}(\vect{\theta})$, which contains the unknown flow function $\vect{\psi}_t(\vect{x}_0)$. 

\section{Discretized consistency training}
The continuous loss can be directly minimized in expectation by sampling the time $t$ from a uniform distribution and by computing the total derivative $\frac{\mathrm{d}}{\mathrm{d}t} \vect{f}_{\vect{\theta}}(\vect{\psi}_t(\vect{x}_0;\vect{x}_1), t)$ by automatic differentiation. However, in practice it is often more convenient to instead optimize a time-discretized loss with a finite difference approximation for the total derivative:
\begin{align} \label{eq: discretized loss}
\begin{split}
    & \mathcal{L}_{\text{disc}}^{\text{cond}}(\vect{\theta}) = \sum_{i=1}^N \lambda(t_i) \mathbb{E}_{\vect{x}_0 \sim p_0(\vect{x}_0),\vect{x}_1 \sim \pi(\vect{x}_1 \mid \vect{x}_0)} \left[|| \Delta \vect{f}_{\vect{\theta}} ||^2\right]~, \\ 
    &~~~ \text{with} \\ 
    & \Delta \vect{f}_{\vect{\theta}} = \vect{f_\theta}(\vect{\psi}_{t_{i+1}}(\vect{x}_0; \vect{x}_1),t_{i+1}) - \vect{f}_{\vect{\theta}^-}(\vect{\psi}_{t_{i}}(\vect{x}_0; \vect{x}_1),t_{i})~.
\end{split}
\end{align}
where $\vect{x}_1 \mid \vect{x}_0$ are sampled according to the noise-coupling $\pi(\vect{x}_1 \mid \vect{x}_0)$. In this expression, $\vect{\theta}^-$ denotes a frozen copy of the parameters which does not require gradients. This loss is in fact unbiased for $\Delta t \rightarrow 0$, as it was shown in \cite{song2023consistency}.
The boundary condition can be enforced through the parametrization introduced in \citep{karras2022elucidating}:
\begin{equation*}
\label{eq:boundary}
    \vect{f_{\theta}}(\vect{x},t) = c_{\mathrm{skip}}(t)\vect{x} + c_{\mathrm{out}}(t)\vect{F_\theta}(\vect{x},t),
\end{equation*}
where $\vect{F_\theta}$ is a neural network and $c_{\mathrm{skip}}$ and $c_{\mathrm{out}}$ are specified such that $c_{\mathrm{skip}}(0)=1$ and $c_{\mathrm{out}}(0)=0$.

\section{Consistency Models with Variational Coupling}
Our method consists in learning a conditional coupling $q_{\vect{\phi}}(\vect{x}_1 \mid \vect{x}_0)$ with a neural network $\vect{g_\phi}(\vect{x})$ parametrized by $\vect{\phi}$, which we refer to as the encoder in the following given its analogy with VAEs. During training, we can sample noise conditionally from $\pi(\vect{x}_1 \mid \vect{x}_0)=q_{\vect{\phi}}(\vect{x}_1 \mid \vect{x}_0)p_0(\vect{x}_0)$ instead of the independent noise commonly used in CT, and obtain noisy states for a given time step $t$ as follows:
\begin{align*}
    \begin{split}
    \vect{x}_t &= \vect{\psi}_t(\vect{x}_0; \vect{x}_1), \\
    \quad \vect{x}_1 \sim q_{\vect{\phi}}(\vect{x}_1 \mid \vect{x}_0) &= \mathcal{N}( \vect{x}_1;\vect{g}_{\vect{\phi}}^{\mu}(\vect{x}_0),~\vect{g}_{\vect{\phi}}^{\sigma}(\vect{x}_0)^2\rmI),
    \end{split}
\end{align*}
where we express the corresponding coupled noise $\vect{x}_1$ using the Gaussian reparameterization formula:
\begin{equation*}
    \vect{x}_1 = \vect{g}_{\vect{\phi}}^{\mu}(\vect{x}_0) + ~\vect{g}_{\vect{\phi}}^{\sigma}(\vect{x}_0) \cdot \vect{\epsilon}, \quad \vect{\epsilon}\sim\mathcal{N}(\vect{\epsilon};\bm{0},\vect{I})
\end{equation*}
Here, we restricted our attention to linear forward models of the form $\vect{\psi}_t(\vect{x}_0; \vect{x}_1) = a_t \vect{x}_0 + b_t \vect{x}_1$, which encompasses most models used in the diffusion and flow-matching literature. Moreover $\vect{g}_{\vect{\phi}}^{\mu}(\vect{x}_0)$ and $\vect{g}_{\vect{\phi}}^{\sigma}(\vect{x}_0)$ denote the mean and scale output of the encoder, which define the signal-noise coupling (see Appendix \ref{app:diag} for a visual representation). Both $\vect{g}_{\vect{\phi}}^{\mu}(\vect{x}_0)$ and $\vect{g}_{\vect{\phi}}^{\sigma}(\vect{x}_0)$ preserve the same dimensionality of the input signal.
The encoder network that produces the coupling distribution $q_{\vect{\phi}}(\vect{x}_1 \mid \vect{x}_0)$ can be trained end-to-end alongside the consistency model, as shown in Algorithm \ref{alg:train-vae}. This results in a joint optimization where the consistency network adjusts its constancy to minimize its total derivative along the trajectories while the encoder implicitly moves the trajectories towards the space of constancy of the model. In fact, the velocity field $\vect{u}_t(\vect{x}_t)$ depends on the coupling, since $\pi(\vect{x}_1 \mid \vect{x}_0)$  affects the expectation in Eq.~\eqref{eq: velocity from conditional}.
\begin{algorithm}[tb]
   \caption{Variational Consistency Training (VCT)}
   \label{alg:train-vae}
\begin{algorithmic}
   \STATE \textbf{Input:} data distribution $p_{\mathrm{data}}$, initial model parameter $\vect{\theta}$, initial encoder parameter $\vect{\phi}$, learning rate $\eta$, EMA rate $\mu$, distance function $d(\cdot,\cdot)$, consistency weighting $\lambda_{\mathrm{ct}}(\cdot)$, KL weighting $\lambda_{\mathrm{KL}}$
   \STATE $\vect{\theta}_{\mathrm{EMA}} \leftarrow \vect{\theta}$, $\vect{\phi}_{\mathrm{EMA}} \leftarrow \vect{\phi}$ and $k \leftarrow 0$
   \REPEAT
   \STATE Sample $\vect{x}_0 \sim p_{\mathrm{data}}$,  $t\sim p(t)$, $r=t-\Delta t$
   \STATE Sample $\vect{\epsilon} \sim N(\rmzr,\rmI)$
   \STATE $\vect{x}_1 \leftarrow \vect{g}_{\vect{\phi}}^{\mu}(\vect{x}_0) + ~\vect{g}_{\vect{\phi}}^{\sigma}(\vect{x}_0) \vect{\epsilon}$
   \STATE $\vect{x}_{t} \leftarrow a_t\vect{x}_0+b_t\vect{x}_1$
   \STATE $\vect{x}_{r} \leftarrow a_r\vect{x}_0+b_r\vect{x}_1$
   \STATE $\mathcal{L}_{\text{disc}}^{\text{cond}}(\vect{\theta, \phi})\leftarrow \lambda_{\mathrm{ct}}(t)d(\vect{f_{\theta}}(\vect{x}_{t}, t),\vect{f_{\theta^-}}(\vect{x}_{r}, r))$
   \STATE $\mathcal{L}_{\mathrm{KL}}(\vect{\phi})\leftarrow \KL(\mathcal{N}(\vect{g}_{\vect{\phi}}^{\mu}(\vect{x}_0),~\vect{g}_{\vect{\phi}}^{\sigma}(\vect{x}_0)^2\rmI) || \mathcal{N}(\rmzr, \rmI))$
   \STATE $\mathcal{\tilde{L}}(\vect{\theta, \phi})\leftarrow \mathcal{L}_{\text{disc}}^{\text{cond}}(\vect{\theta, \phi}) + \lambda_{\mathrm{KL}}\mathcal{L}_{\mathrm{KL}}(\vect{\phi})$
   \STATE $\vect{\theta} \leftarrow \vect{\theta} - \eta \nabla_{\vect{\theta}}\mathcal{\tilde{L}}(\vect{\theta}, \vect{\phi})$
   \STATE $\vect{\phi} \leftarrow \vect{\phi} - \eta \nabla_{\vect{\phi}}\mathcal{\tilde{L}}(\vect{\theta}, \vect{\phi})$
   \STATE $\vect{\theta}_{\mathrm{EMA}} \leftarrow$ \texttt{stopgrad}$(\mu \vect{\theta}_{\mathrm{EMA}} + (1-\mu)\vect{\theta})$
   \STATE $\vect{\phi}_{\mathrm{EMA}} \leftarrow$ \texttt{stopgrad}$(\mu \vect{\phi}_{\mathrm{EMA}} + (1-\mu)\vect{\phi})$
   \STATE $k \leftarrow k + 1$
   \UNTIL convergence
\end{algorithmic}
\end{algorithm}
This formulation results in a viable generative model as long as the noise at time $1$ remains approximately $p_1(\vect{x}_1)$, since severe deviation from the prior induced by the coupling would result in improper initialization for the sampling procedure and consequently in reduced sample quality. We therefore add a KL divergence as a regularizer, $\KL(q_{\vect{\phi}}||p_1)$, resulting in a loss resembling the Evidence Lower Bound loss of Variational Autoencoders \cite{kingma2013auto}:
\begin{align}
    \label{eq:loss}
    \tilde{\mathcal{L}}(\vect{\theta}, \vect{\phi}) &= \mathcal{L}_{\text{disc}}^{\text{cond}}(\vect{\theta}, \vect{\phi}) \\ \nonumber &+ \mean{\KL(q_{\vect{\phi}}(\vect{x}_1 \mid \vect{x}_0)||\mathcal{N}(\vect{x}_1; \rmzr, \rmI))}{\vect{x}_0}.
\end{align}
While using an encoder to learn the data-noise coupling requires additional computation during training, we empirically find that a relatively small encoder is enough to learn an effective coupling, which results only in a minor increase of training time (see Appendix \ref{app:exp}). At sampling, the speed and computational requirements are identical to vanilla CMs for the one-step procedure, while for multistep sampling we need to account for additional forward passes of the encoder as shown in Algorithm \ref{alg:sampling-vae}.

\subsection{Connection with variational autoencoders}

In this section, we will consider the special case with constant unit time weighting $\lambda_{\mathrm{ct}}(t) = 1, \forall t$.
\paragraph{ELBO perspective.}
First, we demonstrate the relationship between our model and VAE in terms of their objective functions. Specifically, our loss function in Eq.~\ref{eq:loss} serves as an upper bound on a standard VAE loss, where the latent vector $\vect{x}_1$ is regularized to be close to the prior $p_1$. Using the triangle inequality, we can establish the following bound for $\mathcal{L}_{\text{disc}}^{\text{cond}}$:
\begin{align} \label{eq: VAE1 loss}
   & \norm{\vect{x}_0 - \vect{f}_{\vect{\theta}}(\vect{x}_1,1)}^2 \leq \\ \nonumber
   &N \sum_{i=0}^N \norm{\vect{f_\theta}(\vect{\psi}_{t_{i+1}}(\vect{x}_0; \vect{x}_1),t_{i+1}) - \vect{f}_{\vect{\theta}^-}(\vect{\psi}_{t_{i}}(\vect{x}_0; \vect{x}_1),t_{i})}^2.
\end{align}
Given that the KL terms in Eq.~\ref{eq:loss} and the VAE loss are identical, our loss function serves as an upper bound on the loss of a VAE with an encoder $(\vect{g}_{\vect{\phi}}^{\mu}(\vect{x}_0),\vect{g}_{\vect{\phi}}^{\sigma}(\vect{x}_0))$ and a prior $\mathcal{N}(\vect{x}_1; \rmzr, \rmI)$. Since the VAE loss represents an evidence lower bound, it follows that the consistency loss also provides a lower bound on the model evidence:
\begin{proposition} The following upper bound for negative log-density holds:
    \begin{align*}
     - \log p_{\theta}(\vect{x}_0)
     &\leq \frac{1}{2\sigma^2}\,\mathbb{E}_{q_{\phi}(\vect{x}_1 \mid \vect{x}_0)} \|\vect{x}_0 - \vect{f_\theta}(\vect{x}_1, 1)\|^2 + \mathcal{L}_{\mathrm{KL}}(\vect{\phi})
     \\&\leq \frac{1}{2\sigma^2} \int_0^1 \mathbb{E} \left[ \left\| \frac{\mathrm{d}}{\mathrm{d}t} \vect{f_\theta}(\psi_t, t) \right\|^2 \right] \mathrm{d}t + \mathcal{L}_{\mathrm{KL}}(\vect{\phi}),
\end{align*}
where $\mathcal{L}_{\mathrm{KL}}(\vect{\phi}):=\KL(q_{\vect{\phi}}(\vect{x}_1 \mid \vect{x}_0)||\mathcal{N}(\vect{x}_1; \rmzr, \rmI))$.
\end{proposition}
This proposition also establishes the connection between the minimization objective of CT-VC and that of the continuous-time CM as $N\rightarrow \infty$ (proof in Appendix \ref{app:elbo}).

Compared to traditional VAEs, our method can be viewed as a time-dependent modification where the transition kernel smoothly interpolates between delta distributions centered at datapoints and a Gaussian distribution.
\paragraph{Varying $\beta$.}
In both our model and VAE, the latent vector needs to approximately follow a normal distribution to avoid deviating from the prior. However, previous studies~\citep{hoffman2016elbo,rosca2018distribution,aneja2021contrastive} have observed that VAE's aggregated posterior fails to match the prior. The same problem could occur in our model without additional tricks (see Sec.~\ref{sec:beta}).
To mitigate this prior-posterior mismatch, we introduce a scalar hyperparameter $\beta$ to control the strength of the KL regularization. This was first introduced by \citet[$\beta$-VAE][]{higgins2022beta} for a different purpose (inducing disentangled latent representation) in the VAE context.
\begin{align}
 \min_{\vect{\theta}, \vect{\phi}}\mathcal{L}_{\text{disc}}^{\text{cond}}(\vect{\theta}, \vect{\phi}) + \beta \mathbb{E}_{\vx_0}[\KL(q_{\vect{\phi}}(\vect{x}_1 \mid \vect{x}_0)||\mathcal{N}(\vect{x}_1; \rmzr, \rmI))].
 \label{eq:beta_loss}
\end{align}
By carefully selecting the value of $\beta$, we can achieve a proper balance between flexibility and proximity to the prior.
To understand the effect of $\beta$ in our model, we present an alternative form of our objective function. Formally, we can view the minimization of Eq.~\ref{eq:beta_loss} as the relaxed Lagrangian problem of the following optimization problem:
\begin{align*}
 &\min_{\vect{\theta}, \vect{\phi}}\mathcal{L}_{\text{disc}}^{\text{cond}}(\vect{\theta}, \vect{\phi}) \\ \nonumber &\mathrm{s.t.} \quad \KL(q_{\vect{\phi}}(\vect{x}_1 \mid \vect{x}_0)||\mathcal{N}(\vect{x}_1; \rmzr, \rmI)) < \delta,
\end{align*}
As $\delta$ approaches 0, the coupling becomes $\pi(\vect{x}_1 \mid \vect{x}_0) =p_0(\vx_0)p_1(\vx_1)$, indicating that our model encompasses the standard CT.
In VAE, selecting appropriate values of $\beta$ to achieve reasonable generation performance is generally challenging. Values too close to zero result in strong deviation from the prior, while extremely large values cause over-smoothed decoders~\citep{takida2022preventing}, leading to blurry samples. However, our model does not suffer from this over-regularization issue. While tuning $\beta$ remains crucial in our model, as demonstrated in Section~\ref{sec:beta}, unlike VAE, increasing values of $\beta$ does not cause the oversmoothing problem but simply reduces our model to the standard CT. Consequently, the CT training objective enables sharp sample generation even when the posterior approximation is nearly a normal distribution.

\begin{algorithm}[tb]
   \caption{Multistep Variational Consistency Sampling}
   \label{alg:sampling-vae}
\begin{algorithmic}
   \STATE \textbf{Input:} Consistency model $\vect{f_\theta}$, encoder $\vect{g_{\phi}}$, sequence of time points $\tau_1 > \tau_2 > \dots > \tau_{N-1}$, initial noise $\vect{\hat{x}}_T$
   \STATE $\vect{x} \leftarrow \vect{f_\theta}(\vect{\hat{x}}_T, T)$
   \FOR{$n=1$ \textbf{to} $N-1$} 
   \STATE Sample $\vect{\epsilon}\sim \mathcal{N}(\rmzr, \rmI)$
   \STATE $\vect{x}_1 \leftarrow \vect{g}_{\vect{\phi}}^{\mu}(\vect{x}) + ~\vect{g}_{\vect{\phi}}^{\sigma}(\vect{x}) \vect{\epsilon}$
    \STATE $\vect{\hat{x}}_{\tau_n} \leftarrow a_{\tau_n}\vect{x} + b_{\tau_n}\vect{x}_1$
   
   \STATE $\vect{x} \leftarrow \vect{f_\theta}(\vect{\hat{x}}_{\tau_n}, \tau_n)$
   \ENDFOR
   \STATE \textbf{Output}: $\vect{x}$
\end{algorithmic}
\end{algorithm}

\subsection{Choosing the $\beta$ parameter}
 The most common weighting function $\lambda_{\mathrm{ct}}(t)$ for CMs is an adaptive weighting scheme that changes over training based on how fine grained the discretization is, or equivalently the distance between consecutive time steps in ECM. At a given training iteration, for two adjacent time steps $t_i$ and $t_{i+1}$, we have:
 \begin{align*}
    \lambda_{\mathrm{ct}}(t_{i+1}) = \frac{1}{\Delta_{t_{i+1}}}, \quad \Delta_{t_{i+1}}=t_{i+1} - t_i
 \end{align*}
 For the models trained with such an adaptive weighting function, we found it hard to tune $\beta$ to a single scalar value. The magnitude of the weights increases during training as the discretization scheme becomes more fine-grained and $\Delta_{t_{i+1}}$ becomes smaller, changing the balance between consistency loss and KL regularization, resulting in a very strong regularization at the early stages of training, or a too weak one at the later stages. A simple yet effective solution is to use an adaptive scaling for the KL regularization that changes according to the discretization scheme. To do so, we take as a reference the weighting of the consistency loss at the last step $t_N=\sigma_{\mathrm{max}}$, and define the adaptive KL weighting as:
\begin{align*}
\lambda_{\mathrm{KL}} = \beta \lambda_{\mathrm{ct}}(t_N) = \frac{\beta}{\Delta_{t_{N}}}.
\end{align*}
This way, we only need to specify the scalar hyperparameter $\beta$, and it will have a consistent regularization strength over training, as it increases whenever the discretization scheme is changed, which reflects in $\Delta_{t_{N}}$ becoming smaller.
For ECM models trained on ImageNet, which use the EDM-style weighting function \begin{align*}
    \lambda_{\mathrm{ct}}(t) = \frac{1}{t^2} + \frac{1}{\sigma_{\mathrm{data}}^2},
\end{align*}
we simply select a fixed $\beta$ scalar for the whole training, as the discretization scheme does not affect the magnitude of the weights.

\section{Experiments}
\label{sec:experiments}
In the following, we show that learning the data-noise coupling with our method is a simple yet effective improvement for CT. We adapt our Variational Coupling (VC) technique to two established baselines: improved Consistency Training (iCT) \citep{songimproved} and Easy Consistency Tuning (ECM) \citep{geng2024consistency}.
 Note that for the latter, the model is initialized with the weights of a pretrained score model from \citep{karras2022elucidating}, while in the former the weights are initialized at random. More details about the baselines are provided in Appendix \ref{app:baselines}. In this work, we consider only the framework of CT, where the unbiased vector field estimator $\vect{u}_t(\vect{x}_t)$ from equation \ref{eq: velocity from conditional} is used to approximate the noisy states $\vect{x}_t$ during training, as opposed to Consistency Distillation that uses a pretrained score model as a teacher. We evaluate the models on the image datasets Fashion-MNIST \citep{xiao2017/online}, CIFAR-10 \citep{krizhevsky2009learning}, FFHQ $64 \times 64$ \citep{karras2019style} and (class-conditional) ImageNet $64 \times 64$ \citep{deng2009imagenet}. To learn the coupling, we add a smaller version of the neural network used for CT, without time conditioning and with weights always initialized at random. For all the models, we use the variance exploding transition kernel (iCT-VE and ECM-VE) used in \citep{karras2022elucidating} and \citep{songimproved}, with $a_t = 1$, $b_t=t$, and the linear interpolation kernel (iCT-LI and ECM-LI) commonly used in Flow Matching \citep{lipmanflow}, with $a_t= 1-t/\sigma_{\mathrm{max}}$ and $b_t= t/\sigma_{\mathrm{max}}$ (details in Appendix \ref{app:fmbc}). For both kernels, we set $\sigma_{\mathrm{min}}=0.002$ and $\sigma_{\mathrm{max}}=80$. We report more experimental details in Appendix \ref{app:exp}, comparison with other models in Appendix \ref{app:fids}, while samples obtained with our best models are shown in Appendix \ref{app:qualitative}.

\subsection{Baselines and models}
As baselines, we re-implement the iCT and ECM models, corresponding to our iCT-VE and ECM-VE. As an additional model, we add CT with the minibatch Optimal Transport Coupling (-OT) proposed in \citep{pooladian2023multisample, tong2023conditional} and used in \citep{issenhuth2024improving}, to compare the effectiveness and scalability of the OT coupling with the learned one. Finally, we combine the baselines with our proposed Variational Coupling (-VC). For the models with learned coupling, we use gradient clipping with a large value (200 in all the experiments) to avoid instabilities at the early stages of training. 

\subsection{Ablation for different $\beta$}
\label{sec:beta}
To see the effect of $\beta$ on the generation performance, we compare the results for different values on the FashionMNIST dataset for iCT-VE-VC in Table \ref{tab:beta}. As expected, with small values of $\beta$, the coupling distribution deviates from the sampling distribution and the performance degenerates, while increasing $\beta$ to high values reduces the benefits of the learned coupling. More detailes on how we tune $\beta$ are reported in Appendix \ref{app:beta}. 

\begin{table}[h]
    \centering
    \resizebox{0.45\columnwidth}{!}{\begin{tabular}{lll}
    \multicolumn{3}{c}{\textbf{FID for different $\beta$}} \\ 
    \toprule
          & 1 step & 2 steps\\
        \midrule
        $\beta=5$ & $12.53$ & $5.69$ \\ 
        \midrule
        $\beta=15$ & $4.97$ & $\bm{2.34}$\\
        \midrule
        $\beta=30$ & $\bm{3.88}$ & $2.37$ \\
        \midrule
        $\beta=60$ & $3.90$ & $2.67$ \\
    \bottomrule        
    \end{tabular}
    }
    \caption{Comparison of FID performance (lower is better) for one and two sampling steps, for varying values of $\beta$. The models are iCT-VE-VC and trained on the FashionMNIST dataset with the same settings described in appendix \ref{app:exp}. Best entries in bold.}
    \label{tab:beta}
\end{table}
\begin{table}[h]
    \centering
\begin{tabular}{c}
    \textbf{1-step / 2-step FID for iCT-based models} \\
\end{tabular}
\vspace{0.5cm}
\resizebox{0.75\columnwidth}{!}{\begin{tabular}{lll}
\toprule
     Model & Fashion-MNIST & CIFAR10\\
\midrule
iCT-VE$^*$ & - & $\bm{2.83}$ / $2.46$ \\ 
\midrule
iCT-VE & $4.79$ / $3.54$ & $3.61$ / $2.79$ \\
\midrule
iCT-LI & $4.75$ / $3.46$ & $3.81$ / $2.87$ \\
\midrule
iCT-VE-OT & $4.42$ / $2.82$ & $3.28$ / $2.66$ \\
\midrule
iCT-LI-OT & $4.41$ / $2.91$ & $3.42$ / $2.77$ \\
\midrule
\rowcolor{gray!15}
iCT-VE-VC (VCT) & $3.88$ / $2.37$ & $2.86$ / $\bm{2.32}$ \\
\midrule
\rowcolor{gray!15}
iCT-LI-VC (VCT) & \bm{$3.62$} / \bm{$2.22$} & $2.94$ / $\bm{2.32}$ \\
\bottomrule  
\end{tabular}
    }
    \caption{Comparison of FID (lower is better, reported as 1-step / 2-step performance) for different models based on iCT. The model marked with a $*$ is the baseline as reported in \citep{songimproved}. All the other models are from our re-implementation. The best entries are highlighted in bold.}
    \label{tab:ict}
\end{table}

\begin{table}[h]
\centering
\small 
\resizebox{\columnwidth}{!}{
\begin{tabular}{llll}
\toprule
Model & CIFAR10 & FFHQ ($64\times 64$) & ImageNet ($64\times 64$) \\ 
\midrule
ECM-VE$^*$ & 3.60 / 2.11 & - & 5.51$^\dagger$ / 3.18$^\dagger$ \\
ECM-VE & 3.68 / 2.14 & 5.99 / 4.39 & 5.26 / 3.22 \\
ECM-LI & 3.65 / 2.14 & 6.42 / 4.73 & 5.13 / 3.20 \\
ECM-VE-OT & 3.46 / 2.13 & 6.11 / 4.68 & 6.02 / 4.27 \\
ECM-LI-OT & 3.49 / 2.13 & 6.19 / 4.73 & 5.63 / 4.09 \\
\rowcolor{gray!15}
ECM-VE-VC (VCT) & \textbf{3.26} / \textbf{2.02} & \textbf{5.47} / \textbf{4.16} & 5.08 / 3.15 \\
\rowcolor{gray!15}
ECM-LI-VC (VCT) & 3.39 / 2.09 & 5.57 / 4.29 & \textbf{4.93} / \textbf{3.07} \\
\bottomrule
\end{tabular}
}
\caption{Comparison of FID (lower is better, reported as 1-step / 2-step performance) for different models based on ECM. The model marked with a $*$ is the baseline as reported in \citep{geng2024consistency}. All the other models are from our re-implementation. The best entries are highlighted in bold. For ImageNet, the results marked with $\dagger$ are obtained with models trained for $100$k iterations, while the others use $200$k iterations. Comprehensive comparisons with additional baselines are presented in Tables~\ref{tab:appcifar} and~\ref{tab:fid_vs_nfe}.}
\label{tab:ecm}
\end{table}
\subsection{Results}
\begin{figure}
    \centering
    \begin{subfigure}[b]{\linewidth}
        \centering
        \includegraphics[width=0.48\linewidth]{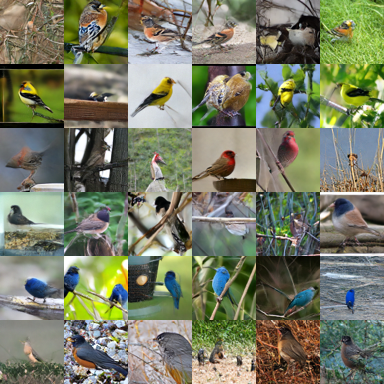}
        \includegraphics[width=0.48\linewidth]{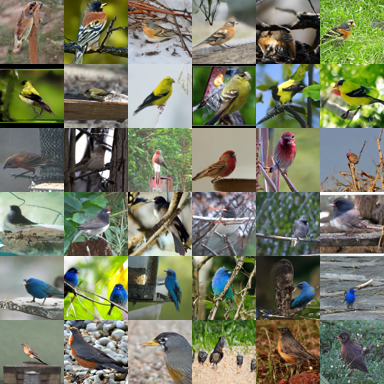}
        \caption{1-step (FID=5.13, left) and 2-step (FID=3.20, right) samples from ECM-LI.}
        \label{fig:top_row}
    \end{subfigure}
    \hfill
    \begin{subfigure}[b]{\linewidth}
        \centering
        \includegraphics[width=0.48\linewidth]{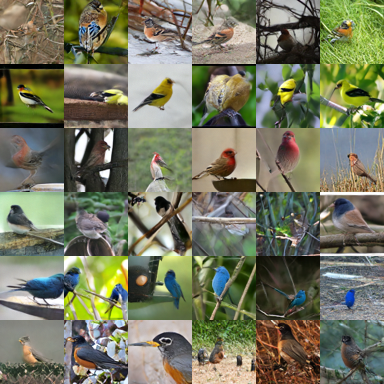}
        \includegraphics[width=0.48\linewidth]{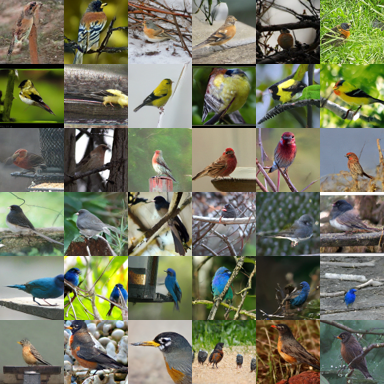}
        \caption{1-step (FID=$4.93$, left) and 2-step (FID=$3.07$, right) samples from ECM-LI-VC.}
        \label{fig:bottom_row}
    \end{subfigure}
    \caption{Visual comparison of generated class-conditional samples on ImageNet $64\times64$.}
    \label{fig:imnet_main1}
\end{figure}
\begin{figure}[ht!]
    \centering
        \centering
        \includegraphics[width=0.8\linewidth]{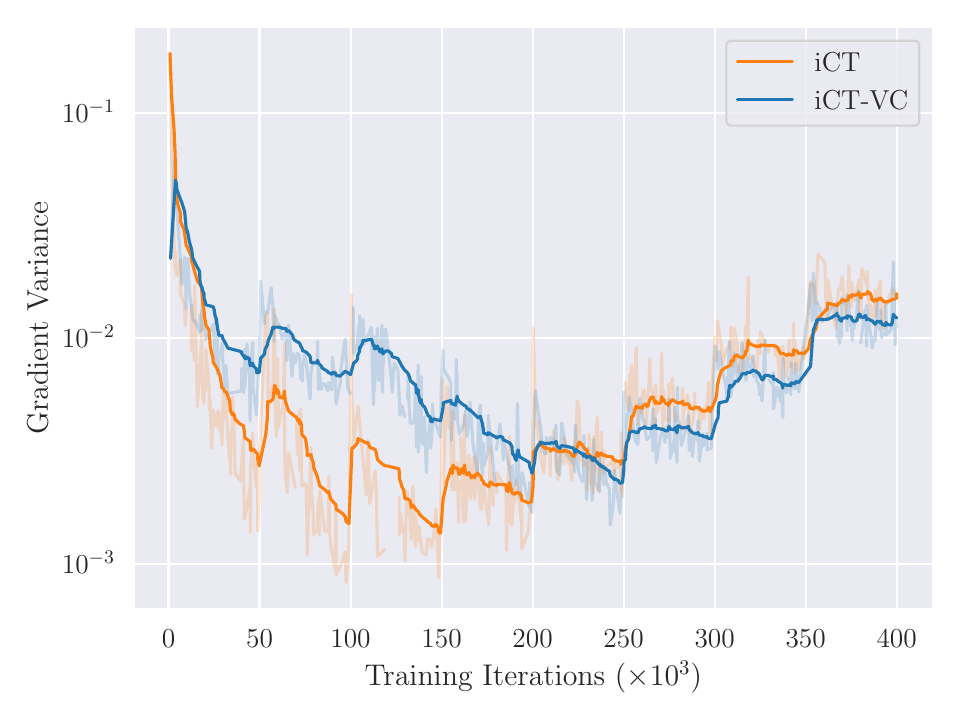}
        \includegraphics[width=0.8\linewidth]{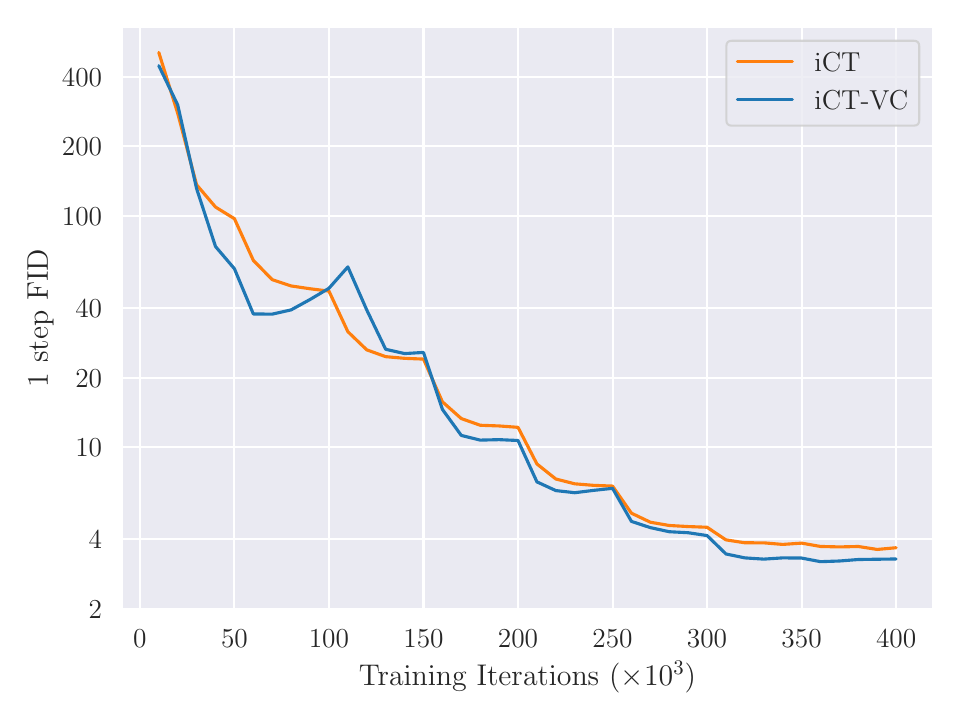}
    \caption{The top graph shows a comparison of gradient variance during training for iCT-VE and iCT-VE-VC on CIFAR10. We plot the variance for each epoch (shaded) and its exponential moving average with smoothing factor 0.9. Especially later during training, the model with learned coupling exhibits lower variance, which results in improved performance, shown in terms of 1-step FID in the bottom graph. For a fair comparison, we did not use gradient clipping for iCT-VC in this run.}
    \label{fig:variance}
\end{figure}
In tables \ref{tab:ict} and \ref{tab:ecm} we report the 1 and 2 step sample quality evaluated with Frechet Inception Distance (FID) \citep{heusel2017gans}, for both the results reported in the original papers and our re-implementations. For high-dimensional data, we only use models based on ECM, as they require lower computational budget, while for FashionMNIST we only use models based on iCT as there is no available pretrained EDM model.\\
\textbf{FashionMNIST}: We choose FashionMNIST as a first benchmark to test the performance of iCT. On this dataset, we use a small version of DDPM++, with $64$ model channels instead of $128$ and no attention, and batch size $128$. Our variant with Variational Coupling outperforms both iCT and iCT-OT, with best performance obtained with the LI transition kernel, showing the benefit of the learned coupling. \\
\textbf{CIFAR-10}: For all the CIFAR-10 experiments, we use the DDPM++ architecture from \citep{songscore} as implemented in \citep{karras2022elucidating}, with EMA rate $0.9999$ as in \citep{geng2024consistency}. While this differs from the settings in \citep{songimproved}, we found it to work better in our re-implementation. The remaining hyperparameters are the same as used in the respective baselines. From the results, we can see how using the learned coupling results in improved performance for both one and two steps generation, outperforming all the re-implemented baselines.
The 1-step result from the original iCT is superior to our model. However, to the best of our knowledge, there is no open-source implementation that can reproduce the results reported in the paper. The learned coupling outperforms the minibatch OT coupling in all cases, as it is less affected by the effective (per device) batch size and the data dimensionality. Finally, our 2-step sampling performance for ECM-VE-VC is on par with the current SoTA achieved by other methods with similar settings \citep{wang2024stable, lee2024truncated, lu2024simplifying}.
We empirically compare the variance of the gradients for iCT-VE and iCT-VE-VC, and show in Figure \ref{fig:variance} how the resulting reduced variance corresponds to improved FID score. In particular, in early training the model with VC exhibits higher variance, which can be attributed to the fact that the encoder is still learning the coupling. As training continues, the coupling becomes effective at providing better data-noise pairs to the model, which results in reduced gradient variance and improved generative performance.\\
\textbf{FFHQ $64 \times 64$}: We use FFHQ $64 \times 64$ as an additional dataset to assess our method on higher-dimensional data. We reuse the same training settings used for CIFAR10, without additional tuning, and with the same network architecture used in EDM. While the results are worse than current SoTA generative models (e.g. 2.39 FID from EDM), they confirm the benefit of using the learned coupling over the baselines. Moreover, the results highlight the limits of using the minibatch OT coupling, which scales poorly with increased data dimensionality and in some cases performs worse than the independent coupling.\\
\textbf{ImageNet $64 \times 64$} (class conditional): As a baseline, we reuse the settings from ECM with the EDM2-S architecture and batch size $128$. While the baseline is trained for $100k$ iteration, we found that our models with Variational Coupling needed more time to converge properly, as the encoder weights are not pretrained and initialized at random. We therefore train our re-implemented baselines and models for $200k$ iterations instead, while we report the performance of our models trained for $100k$ iterations in Appendix \ref{app:imgnet100k}. In the $200k$ case, the models with Variational Coupling outperform the other models, with the LI kernel obtaining the best overall FID, while the OT coupling performs poorly due to the small batch size and high data dimensionality. In Figure \ref{fig:imnet_main1} we compare samples from ECM-LI and ECM-LI-VC, where we can see how the images generated with VC are more clear and detailed.

\section{Conclusions}
In this work, we introduced a novel approach to Consistency Training (CT) by incorporating a variational noise coupling mechanism. Our method leverages an encoder-based coupling function to learn a data-dependent noise distribution, which results in improved generative performance. By framing CT within the Flow Matching perspective, we provided a principled way to introduce adaptive noise coupling while maintaining the efficiency of standard CT. Empirical results on multiple image benchmarks, demonstrate that our approach consistently outperforms baselines in one and two-step generation settings. Our findings highlight the potential of learned coupling in CT and suggest several promising directions for future work. These include exploring more expressive posterior distributions, extending our method to the continuous-time CT formulation, and integrating variational consistency training with other recent CT improvements. We hope this work contributes to the broader understanding of the effect of coupling in CT and inspires further advancements in efficient generative sampling techniques.
\newpage
\section*{Impact Statement}

This paper presents work whose goal is to advance the field of 
Machine Learning. There are many potential societal consequences 
of our work, none which we feel must be specifically highlighted here.

\section*{Acknowledgment}
We sincerely thank the reviewers for their constructive feedback, which greatly helped improve this
work. We are also deeply grateful to our colleague, Naoki Murata, for revising the manuscript and
providing valuable comments.

\bibliography{citations}
\bibliographystyle{icml2025}

\newpage
\appendix
\onecolumn

\section{Consistency Models with Linear Interpolation Kernel}
\label{app:fmbc}
In addition to the variance exploding forward process commonly used in CT, here we propose to use the linear interpolation kernel commonly used in Flow Matching:
\begin{align}
    \vect{x}_t = (1-t)\vect{x}_0 + t \vect{x}_1.
\end{align}
We reuse all of the building blocks from iCT and ECM and make only the necessary adjustments. Accounting for the boundary conditions, the transition kernel becomes: 

\begin{align}
   \vect{x}_t = \left( 1-\frac{t}{\sigma_{\mathrm{max}}}\right)\vect{x}_0 + \left(\frac{t}{\sigma_{\mathrm{max}}}\right) \vect{x}_1 \sigma_{\mathrm{max}}. 
\end{align}

Other crucial components for stability during training are the scaling factors $c_{\mathrm{in}}$, $c_{\mathrm{skip}}$ and $c_{\mathrm{out}}$, and we derive them for the linear interpolation kernel following the same procedure used in \citep{karras2022elucidating}, also accounting for the boundary conditions when $\sigma_{\mathrm{min}}\neq0$:

\begin{align}
    c_{\mathrm{in}}(\sigma) &= \cfrac{1}{\sqrt{\sigma_{\mathrm{data}}^2  (1-\frac{\sigma}{\sigma_{\mathrm{max}}})^2 + \sigma^2}} \\
    c_{\mathrm{skip}}(\sigma) &= \cfrac{\sigma_{\mathrm{data}}^2 (1-\frac{\sigma - \sigma_{\mathrm{min}}}{\sigma_{\mathrm{max}} - \sigma_{\mathrm{min}}})}{(\sigma - \sigma_{\mathrm{min}})^2 + \sigma_{\mathrm{data}}^2 (1-\frac{\sigma - \sigma_{\mathrm{min}}}{\sigma_{\mathrm{max}} - \sigma_{\mathrm{min}}})^2} \\
    c_{\mathrm{out}}(\sigma) &= (\sigma - \sigma_{\mathrm{min}})  \sigma_{\mathrm{data}} c_{\mathrm{in}}(\sigma)
\end{align}

\subsection{Derivations}

We report the derivations for the scaling factors used for the linear interpolation transition kernel. We follow the same derivations from \citep{karras2022elucidating} (appendix B.6), where the score matching objective is written as:
\begin{align}
    E||D_{\vect{\theta}} (\vect{y} + \vect{n}; \sigma) - \vect{y} ||_2^2
\end{align}
Where $\vect{y}$ is data sampled from the data distribution with standard deviation $\sigma_\mathrm{data}$ and $\vect{n}$ is a sample from noise distribution with standard deviation $\sigma$. Given this objective, they propose to derive the scaling factors $c_{\mathrm{in}}(\sigma)$, $c_{\mathrm{skip}}(\sigma)$, $c_{\mathrm{out}}(\sigma)$ as follows:
\begin{align}
c_{\mathrm{in}}(\sigma) &= \frac{1}{\sqrt{\mathrm{Var}_{\vect{y},\vect{n}}[\vect{y}+\vect{n}]}} \\
c_{\mathrm{out}}(\sigma)^2 & = \mathrm{Var}_{\vect{y},\vect{n}}[\vect{y} - c_{\mathrm{skip}}(\sigma)(\vect{y}+\vect{n})] \\
c_{\mathrm{skip}}(\sigma) & = \mathrm{arg min}_{c_{\mathrm{skip}}(\sigma)}c_{\mathrm{out}}(\sigma)^2.
\end{align}

In our formulation, we only need to rescale $\vect{y}$ by $1 - \frac{\sigma}{\sigma_{\mathrm{max}}}$ and perform the same derivations. For simplicity, we define $\alpha = 1 - \frac{\sigma}{\sigma_{\mathrm{max}}}$ (omitting the dependence on $\sigma$), and proceed as follows:
\begin{align}
c_{\mathrm{in}}(\sigma) &= \frac{1}{\sqrt{\mathrm{Var}_{\vect{y},\vect{n}}[\alpha \vect{y}+\vect{n}]}} \\
c_{\mathrm{out}}(\sigma)^2 & = \mathrm{Var}_{\vect{y},\vect{n}}[\vect{y} - c_{\mathrm{skip}}(\sigma)(\alpha \vect{y}+\vect{n})] \\
c_{\mathrm{skip}}(\sigma) & = \mathrm{arg min}_{c_{\mathrm{skip}}(\sigma)}c_{\mathrm{out}}(\sigma)^2.
\end{align}

The factor $c_{\mathrm{in}}(\sigma)$ simply becomes:
\begin{align}
c_{\mathrm{in}}(\sigma) &= \frac{1}{\sqrt{\sigma_\mathrm{data}^2 * \alpha^2 + \sigma^2}}.
\end{align}

To derive $c_{\mathrm{out}}(\sigma)$ we can proceed as:
\begin{align}
    c_{\mathrm{out}}(\sigma)^2 & = \mathrm{Var}_{\vect{y},\vect{n}}[\vect{y} - c_{\mathrm{skip}}(\sigma)(\alpha \vect{y}+\vect{n})] \\
    c_{\mathrm{out}}(\sigma)^2 & = \mathrm{Var}_{\vect{y},\vect{n}}[(1 - \alpha c_{\mathrm{skip}}(\sigma))\vect{y} + c_{\mathrm{skip}}(\sigma)\vect{n}] \\
    c_{\mathrm{out}}(\sigma)^2 & = (1 - \alpha c_{\mathrm{skip}}(\sigma))^2\sigma_\mathrm{data}^2 + c_{\mathrm{skip}}(\sigma)^2\sigma^2.
\end{align}

We can use this result to solve for $c_\mathrm{skip}(\sigma)$:
\begin{align}
    0 &= d[c_{\mathrm{out}}(\sigma)^2] / dc_{\mathrm{skip}}(\sigma) \\
    0 &= d[(1 - \alpha c_{\mathrm{skip}}(\sigma))^2\sigma_\mathrm{data}^2 + c_{\mathrm{skip}}(\sigma)^2\sigma^2] / dc_{\mathrm{skip}}(\sigma) \\
    0 &= \sigma_\mathrm{data}^2d[(1 - \alpha c_{\mathrm{skip}}(\sigma))^2]/ dc_{\mathrm{skip}}(\sigma) + \sigma^2d[c_{\mathrm{skip}}(\sigma)^2]/dc_{\mathrm{skip}}(\sigma) \\
    0 &= \sigma_\mathrm{data}^2[2\alpha ^2c_{\mathrm{skip}}(\sigma)-2\alpha] + \sigma^2[2c_{\mathrm{skip}}(\sigma)] \\
    0 &= (\sigma^2 + \alpha ^2\sigma_\mathrm{data}^2)c_{\mathrm{skip}}(\sigma) - \alpha \sigma_\mathrm{data}^2 \\
    c_{\mathrm{skip}}(\sigma) &= \alpha \sigma_\mathrm{data}^2/(\sigma^2 + \alpha^2\sigma_\mathrm{data}^2).
\end{align}

Finally, we can compute $c_\mathrm{out}(\sigma)$:
\begin{align}
    c_{\mathrm{out}}(\sigma)^2 &= (1 - \alpha c_{\mathrm{skip}}(\sigma))^2\sigma_\mathrm{data}^2 + c_{\mathrm{skip}}(\sigma)^2\sigma^2 \\
    c_{\mathrm{out}}(\sigma)^2 &= \left(1 - \left[\frac{\alpha^2\sigma_\mathrm{data}^2}{(\sigma^2 + \alpha^2\sigma_\mathrm{data}^2)}\right]\right)^2\sigma_\mathrm{data}^2 +\left[\frac{\alpha\sigma_\mathrm{data}^2}{(\sigma^2 + \alpha^2\sigma_\mathrm{data}^2)}\right]^2 \sigma^2 \\
    c_{\mathrm{out}}(\sigma)^2 &= \left[\frac{\sigma^2 \sigma_\mathrm{data}}{(\sigma^2 + \alpha^2\sigma_\mathrm{data}^2)}\right]^2 +\left[\frac{\alpha\sigma_\mathrm{data}^2 + \sigma}{(\sigma^2 + \alpha^2\sigma_\mathrm{data}^2)}\right]^2 \\
    c_{\mathrm{out}}(\sigma)^2 &= \frac{(\sigma^2 \sigma_\mathrm{data})^2 + (\sigma \alpha\sigma_\mathrm{data}^2)^2}{(\sigma^2 + \alpha^2\sigma_\mathrm{data}^2)^2} \\
    c_{\mathrm{out}}(\sigma)^2 &= \frac{(\sigma \sigma_\mathrm{data})^2 + (\alpha^2\sigma_\mathrm{data}^2 + \sigma^2)}{(\sigma^2 + \alpha^2\sigma_\mathrm{data}^2)^2} \\
    c_{\mathrm{out}}(\sigma)^2 &= \frac{(\sigma \sigma_\mathrm{data})^2}{(\sigma^2 + \alpha^2\sigma_\mathrm{data}^2)} \\
    c_{\mathrm{out}}(\sigma) &= \frac{\sigma \sigma_\mathrm{data}}{\sqrt{\sigma^2 + \alpha^2\sigma_\mathrm{data}^2}}.
\end{align}

If we want to use the boundary conditions for $\sigma_{\mathrm{min}} \neq 0$, then we can modify $c_\mathrm{skip}(\sigma)$ and  $c_\mathrm{out}(\sigma)$ as:
\begin{align}
     c_{\mathrm{skip}}(\sigma) &= \cfrac{\sigma_{\mathrm{data}}^2  (1-\frac{\sigma - \sigma_{\mathrm{min}}}{\sigma_{\mathrm{max}} - \sigma_{\mathrm{min}}})}{(\sigma - \sigma_{\mathrm{min}})^2 + \sigma_{\mathrm{data}}^2 (1-\frac{\sigma - \sigma_{\mathrm{min}}}{\sigma_{\mathrm{max}} - \sigma_{\mathrm{min}}})^2} \\
    c_{\mathrm{out}}(\sigma) &= (\sigma - \sigma_{\mathrm{min}}) \sigma_{\mathrm{data}} c_{\mathrm{in}}(\sigma),
\end{align}

which satisfy the condition $c_\mathrm{skip}(\sigma_{\mathrm{min}})=1$ and $c_\mathrm{out}(\sigma_{\mathrm{min}})=0$.

\section{Consistency Lower Bound}
\label{app:elbo}
\begin{proposition} The following upper bound for negative log-density holds:
    \begin{align*}
     - \log p_{\theta}(\vect{x}_0) &\leq \frac{1}{2\sigma^2}\,\mathbb{E}_{q_{\phi}(\vect{x}_1 \mid \vect{x}_0)} \|\vect{x}_0 - \vect{f_\theta}(\vect{x}_1, 1)\|^2 + \mathcal{D}_{\mathrm{KL}}\Bigl(q_{\phi}(\vect{z} \mid \vect{x}_0) \,\Big\|\, p(\vect{z})\Bigr) 
     \\&\leq \frac{1}{2\sigma^2} \int_0^1 \mathbb{E} \left[ \left\| \frac{d}{dt} \vect{f_\theta}(\psi_t, t) \right\|^2 \right] dt + \mathcal{D}_{\mathrm{KL}}\Bigl(q_{\phi}(\vect{z} \mid \vect{x}_0) \,\Big\|\, p(\vect{z})\Bigr). 
\end{align*}
\end{proposition}
This proposition establishes the connection between the minimization objective of CT-VC and that of the continuous-time CM.

\begin{proof}
For the triangle inequality, we have:
    \begin{align}
        & \norm{\vect{x}_0 - \vect{f}_{\vect{\theta}}(\vect{x}_1,1)} \leq \\ \nonumber
   & \sum_{i=0}^N \norm{\vect{f_\theta}(\vect{\psi}_{t_{i+1}}(\vect{x}_0; \vect{x}_1),t_{i+1}) - \vect{f}_{\vect{\theta}^-}(\vect{\psi}_{t_{i}}(\vect{x}_0; \vect{x}_1),t_{i})}.
    \end{align}
    We can now square both sides, obtaining:
    
    \begin{align}
    \label{ineq1}
        & \norm{\vect{x}_0 - \vect{f}_{\vect{\theta}}(\vect{x}_1,1)}^2 \leq \\ \nonumber
   & \left(\sum_{i=0}^N \norm{\vect{f_\theta}(\vect{\psi}_{t_{i+1}}(\vect{x}_0; \vect{x}_1),t_{i+1}) - \vect{f}_{\vect{\theta}^-}(\vect{\psi}_{t_{i}}(\vect{x}_0; \vect{x}_1),t_{i})}\right)^2.
    \end{align}
    Now, for the Cauchy-Schwarz Inequality, we can write the right hand side as:
    \begin{align}
    \label{ineq2}
        &\left(\sum_{i=0}^N \norm{\vect{f_\theta}(\vect{\psi}_{t_{i+1}}(\vect{x}_0; \vect{x}_1),t_{i+1}) - \vect{f}_{\vect{\theta}^-}(\vect{\psi}_{t_{i}}(\vect{x}_0; \vect{x}_1),t_{i})}\right)^2 \leq \\
        \nonumber & N\sum_{i=0}^N \norm{\vect{f_\theta}(\vect{\psi}_{t_{i+1}}(\vect{x}_0; \vect{x}_1),t_{i+1}) - \vect{f}_{\vect{\theta}^-}(\vect{\psi}_{t_{i}}(\vect{x}_0; \vect{x}_1),t_{i})}^2.
    \end{align}

Since:
\begin{align}
\log p_{\theta}(\vect{x}_0 \mid \vect{x}_1) \propto -\frac{1}{2\sigma^2}\|\vect{x}_0 - \vect{f_{\theta}}(\vect{x}_1,1)\|^2,
\end{align}
we can write a lower bound on the log density as:

\begin{align}
    \log p_{\theta}(\vect{x}_0) \ge -\frac{1}{2\sigma^2}\,\mathbb{E}_{q_{\phi}(\vect{x}_1 \mid \vect{x}_0)} \|\vect{x}_0 - \vect{f_\theta}(\vect{x}_1,1)\|^2 - \mathcal{D}_{\mathrm{KL}}\Bigl(q_{\phi}(\vect{x}_1 \mid \vect{x}_0) \,\Big\|\, p(\vect{x}_1)\Bigr).
\end{align}

In summary, the ELBO bound for the Gaussian VAE is given by

\begin{align}
 - \log p_{\theta}(\vect{x}_0) \leq \frac{1}{2\sigma^2}\,\mathbb{E}_{q_{\phi}(\vect{x}_1 \mid \vect{x}_0)} \|\vect{x}_0 - \vect{f_\theta}(\vect{x}_1, 1)\|^2 + \mathcal{D}_{\mathrm{KL}}\Bigl(q_{\phi}(\vect{x}_1 \mid \vect{x}_0) \,\Big\|\, p(\vect{x}_1)\Bigr):=L(\vect{x}_0; \theta, \phi).
\end{align}
Combining the inequalities we derived in Eq. \ref{ineq1} and \ref{ineq2}, below we consider the case when $N\rightarrow \infty$. First, we define:
\begin{equation}
    S_N = N\sum_{i=0}^{N} \mathbb{E}_{q(x_1 \mid x_0), p(t_i)} \left[ \| \vect{f_\theta}(\psi_{t_{i+1}}(x_0; x_1), t_{i+1}) - \vect{f_\theta}(\psi_{t_i}(x_0; x_1), t_i) \|^2 \right].
\end{equation}

We assume that $t_i$ is a partition of the interval $[0,1]$ of:
\begin{equation}
    \Delta t := t_{i+1} - t_i = \mathcal{O}\left(\frac{1}{N}\right).
\end{equation}
For small $\Delta t = \mathcal{O}\left(\frac{1}{N}\right)$, we approximate the squared difference in function values using a first-order Taylor expansion\footnote{Here, we assume $\psi_{t_{i+1}}-\psi_{t_{i}}=\mathcal{O}(t_{i+1}-t_i)$}:
\begin{equation}
    \| \vect{f_\theta}(\psi_{t_{i+1}}, t_{i+1}) - \vect{f_\theta}(\psi_{t_i}, t_i) \|^2 \approx \left\| \frac{d}{dt} \vect{f_\theta}(\psi_t, t) \bigg|_{t=t_i} \right\|^2 \cdot \Delta t^2.
\end{equation}

Since $\Delta t = \mathcal{O}\left(\frac{1}{N}\right)$, we substitute:
\begin{equation}
    \| \vect{f_\theta}(\psi_{t_{i+1}}, t_{i+1}) - \vect{f_\theta}(\psi_{t_i}, t_i) \|^2 \approx \frac{1}{N^2} \left\| \frac{d}{dt} \vect{f_\theta}(\psi_t, t) \right\|^2.
\end{equation}
Multiplying by $N$:
\begin{equation}
    S_N=N\sum_{i=0}^{N} \mathbb{E} \left[ \| \vect{f_\theta}(\psi_{t_{i+1}}, t_{i+1}) - \vect{f_\theta}(\psi_{t_i}, t_i) \|^2 \right] \approx \frac{1}{N} \sum_{i=0}^{N} \mathbb{E} \left[  \left\| \frac{d}{dt} \vect{f_\theta}(\psi_t, t) \right\|^2 \right].
\end{equation}

As $N\rightarrow \infty$,
\begin{equation}
    \frac{1}{N} \sum_{i=0}^{N} \mathbb{E} \left[  \left\| \frac{d}{dt} \vect{f_\theta}(\psi_t, t) \right\|^2 \right] \rightarrow  \int_0^1 \mathbb{E} \left[ \left\| \frac{d}{dt} \vect{f_\theta}(\psi_t, t) \right\|^2 \right] dt.
\end{equation}

Thus, multiplying by $N$, we obtain:
\begin{equation}
    \lim_{N \to \infty} S_N = \int_0^1 \mathbb{E} \left[ \left\| \frac{d}{dt} \vect{f_\theta}(\psi_t, t) \right\|^2 \right] dt.
\end{equation}

Combining the above limit with the ELBO bound:
\begin{align}
     - \log p_{\theta}(\vect{x}_0) &\leq \frac{1}{2\sigma^2}\,\mathbb{E}_{q_{\phi}(\vect{x}_1 \mid \vect{x}_0)} \|\vect{x}_0 - \vect{f_\theta}(\vect{x}_1, 1)\|^2 + \mathcal{D}_{\mathrm{KL}}\Bigl(q_{\phi}(\vect{z} \mid \vect{x}_0) \,\Big\|\, p(\vect{z})\Bigr) \label{eq:elbo-vct-1}
     \\&\leq \frac{1}{2\sigma^2} \frac{1}{N} \sum_{i=0}^{N} \mathbb{E} \left[  \left\| \frac{d}{dt} \vect{f_\theta}(\psi_t, t) \right\|^2 \right]  + \mathcal{D}_{\mathrm{KL}}\Bigl(q_{\phi}(\vect{z} \mid \vect{x}_0) \,\Big\|\, p(\vect{z})\Bigr). \label{eq:elbo-vct-2}
     \\&= \frac{1}{2\sigma^2} \int_0^1 \mathbb{E} \left[ \left\| \frac{d}{dt} \vect{f_\theta}(\psi_t, t) \right\|^2 \right] dt + \mathcal{D}_{\mathrm{KL}}\Bigl(q_{\phi}(\vect{z} \mid \vect{x}_0) \,\Big\|\, p(\vect{z})\Bigr), \quad \text{as } N\rightarrow\infty. \label{eq:elbo-vct-3}
\end{align}

\end{proof}
\section{Experimental details}
\label{app:exp}
\subsection{Baselines}
\label{app:baselines}
Here we recap the detials about the two baselines used in this work, the Improved Consistency Training from \citep{songimproved} and Easy Consistency Models from \citep{geng2024consistency}.

\textbf{iCT}: the training procedure uses a discretization of time steps between two values $\sigma_{\mathrm{min}}=0.002$ and $\sigma_{\mathrm{max}}=80$, with the equation from \citep{karras2022elucidating}:
\begin{align}
    \sigma_i = \left( \sigma_{\min}^{1/\rho} + \frac{i-1}{N(k)-1} \left( \sigma_{\max}^{1/\rho} - \sigma_{\min}^{1/\rho} \right) \right)^\rho, \text{ where } i \in [[1, N(k)]],
\end{align}
where $\rho=7$ and $N(k)$ is a scheduler that defines the number of discretization steps at the $k$-th training iteration. $N(k)$ is chosen to be an exponential schedule which starts from $s_0=10$ steps and reaches $s_1=1280$ steps at the end of the training, and is defined as:
\begin{align}
    N(k) = \min(2^{\lceil k / K' \rceil}, s_1) + 1, \quad K' = \left\lceil \frac{K}{\log_2(s_1 / s_0) + 1} \right\rceil.
\end{align}
During training, time steps $t_i$ (or equivalently $\sigma_i$) are sampled following a discrete lognormal distribution:
\begin{align}
    p(\sigma_{i}) \propto \text{erf}\left(\frac{\log(\sigma_{i+1}) - P_{\text{mean}}}{\sqrt{2}P_{\text{std}}}\right) - \text{erf}\left(\frac{\log(\sigma_{i}) - P_{\text{mean}}}{\sqrt{2}P_{\text{std}}}\right),
\end{align}
with $P_{\text{mean}}=-1.1$ and $P_{\text{std}}=2.0$. Then, the steps $t_i$ and $t_{i+1}$ are used in the loss:
\begin{align}
    \mathcal{L}_{\mathrm{ct}}(\vect{\theta, \phi})\leftarrow \lambda_{\mathrm{ct}}(t_i)d(\vect{f_{\theta}}(\vect{x}_{t_{i+1}}, t+1),\vect{f_{\theta^-}}(\vect{x}_{t_i}, t_i)),
\end{align}
whith the time dependent weighting function $\lambda_{\mathrm{ct}}(t_i)=\frac{1}{t_{i+1}-{t_i}}$, and $d(.,.)$ is the Pseudo-Huber loss:
\begin{align}
    d(\vect{x},\vect{y}) = \sqrt{||\vect{x}-\vect{y}||_{2}^{2}+c^{2}} - c.
\end{align}
\\
\textbf{ECM}: ECM aims to simplify and improve the training procedure from iCT. We report here the main differences. Instead of using a discretized grid of time steps, it samples time steps $t$ from a continuous lognormal distribution with  $P_{\text{mean}}=-1.1$ and $P_{\text{std}}=2.0$ ($-0.8$ and $1.6$ for ImageNet). The second time step $r$ used in the discretized training objective is then obtained with a mapping function 
\begin{align}
    p(r|t, \text{iters}) = 1 - \frac{1}{q^a}n(t) = 1 - \frac{1}{q^{\lceil \text{iters}/d \rceil}}n(t),
\end{align}
where $n(t) = 1 + k\sigma(-bt) = 1 + \frac{k} {1+e^{bt}}$, $\sigma(.)$ is the sigmoid function, \textit{iter} is the current training iteration, $k=8$, $b=1$, and $q=2$ for all the models but ImageNet, where $q=4$. The discretization step is made smaller for eight times over training (four times for ImageNet). The loss function is a generalization of the Pseudo-Huber loss, which consists of the L2 loss and an adaptive weighting function $w(\Delta)$. The models are initialized with the weights of pretrained diffusion models, which is shown to greatly improve stability during training and generation performance.

\subsection{Training details}

We report the training details for our models in Tables \ref{app:tab_ict} and \ref{app:tab_ecm}. Note that the baselines are the ones from our reimplementation. The models have the same number of parameters and training hyperparameters regardless of the transition kernel used. In the following, we report additional information important for reproducing out experiments:

\textbf{ECM-LI}: In ECM, the time steps $t$ ar sampled from a lognormal distribution, as done in \citep{karras2022elucidating}. This means that time steps $t>\sigma_{\mathrm{max}}$ can be sampled during training. While this works well when using the variance exploding Kernel, in the linear interpolantion case the time step $t$ cannot exceed $\sigma_{\mathrm{max}}$, and we therefore clip $t$ to be at most $\sigma_{\mathrm{max}}$.

\textbf{Random seeds}: All the training runs are initialized with random seed 42. For sampling and FID computation, we always set the random seed to 32, which was randomly chosen. This differs from what commonly done in EDM, where three different seeds are used to evaluate FID and the best result is reported. While our evaluation can lead to slightly worse results, the evaluation is consistent between our models and reimplemented baselines.

\textbf{2-steps generation}: Like in the original iCT baseline, all the models use $t=0.821$ for CIFAR10 and all the other datasets but ImageNet, where $t=1.526$ is used insetad.

\textbf{Data augmentation}: We scale all the images to have values between $-1$ and $1$. For CIFAR10 we apply random horizontal flip with $50\%$ probability.

\textbf{Differences for ImageNet}: The training procedure for ECM on ImageNet differs slightly from the one for the other datasets. The Adam optimizer is used instead of RAdam, with betas$=(0.9,0.99)$, and inverse square root learning rate decay defined as a function of the current training iteration $i$:
\begin{align}
    \alpha(i) = \frac{\alpha_{\text{ref}}}{\sqrt{\max(i/i_{\text{ref}},1)}},
\end{align}
with $\alpha_{\text{ref}}=0.001$ (the initial learning rate) and $i_{\text{ref}}=2000$ iterations. The Exponential Moving Average uses the power function averaging profile introduced in \citep{karras2024analyzing}. In ECM, three different EMA profiles are tracked during training, with rates $0.01$, $0.05$, and $0.1$. In our reimplementation, we only use the rate $0.1$. The number of times in which the discretization interval changes is reduced from $8$ to $4$, and the loss constant $c$ is set to $0.06$.

\begin{table}[h!]
\centering
\resizebox{0.6\textwidth}{!}{%
\begin{tabular}{lcc}
\textbf{Model Setups} & \textbf{FashionMNIST} & \textbf{CIFAR10} \\
\hline
Model Architecture & DDPM++ & DDPM++\\
Model Channels & 64 & 128 \\
N$^\circ$ of ResBlocks & 4 & 4 \\
Attention Resolution & - & 16 \\
Channel multiplyer & $[2, 2, 2]$ & $[2, 2, 2]$ \\
Model capacity & 13.6M & 55.7M\\
\hline
\textbf{Training Details} & & \\
Minibatch size & 128 & 1024\\
Batch per device & 128 & 512\\
Iterations & 400k & 400k\\
Dropout probability & 30\% & 30\% \\
Optimizer & RAdam & RAdam \\
Learning rate & 0.0001 & 0.0001 \\
EMA rate & 0.9999 & 0.9999 \\
\hline
\textbf{Training Cost} & & \\
Number of GPUs & 1 & 2\\
GPU types & H100 & H100\\
Training time (hours) & 28 & 92 \\
Training time with OT (hours) & 29 & 95 \\
\hline
\textbf{Encoder Details} & & \\
Model Architecture & DDPM++ & DDPM++ \\
Model Channels & 32 & 32 \\
N$^\circ$ of ResBlocks & 1 & 1 \\
Attention Resolution & - & 16 \\
Channel multiplyer & $[2, 2, 2]$ & $[2, 2, 2]$ \\
$\beta$ regularizer & 30 & 30 \\
Encoder Params & 1.5M & 1.6M\\
Training time with Encoder (hours) & 34 & 102 \\
\end{tabular}%
}
\caption{Model Configurations and Training Details for iCT on FashionMNIST and CIFAR10}
\label{app:tab_ict}
\end{table}

\begin{table}[h!]
\centering
\resizebox{0.8\textwidth}{!}{%
\begin{tabular}{lccc}
\textbf{Model Setups} & \textbf{CIFAR10} & \textbf{FFHQ $64\times46$} & \textbf{ImageNet $64 \times 64$}\\
\hline
Model Architecture & DDPM++ & DDPM++ & EDM2-S\\
Model Channels & 64 & 128 & 192 \\
N$^\circ$ of ResBlocks & 4 & 4 & 3\\
Attention Resolution & [16] & [16] & [16, 8] \\
Channel multiplyer & $[2, 2, 2]$ & $[1, 2, 2, 2]$ & $[1, 2, 3, 4]$\\
Model capacity & 55.7M & 61.8M & 280M\\
\hline
\textbf{Training Details} & & \\
Minibatch size & 128 & 128 & 128\\
Batch per device & 128 & 128 & 128\\
Iterations & 400k & 400k & 200k\\
Dropout probability & 20\% & 20\% & 40\% (res $\leq16$)\\
Optimizer & RAdam & RAdam & Adam \\
Learning rate & 0.0001 & 0.0001 & 0.001 \\
EMA rate & 0.9999 & 0.9999 & 0.1 \\
\hline
\textbf{Training Cost} & & \\
Number of GPUs & 1 & 1 & 1\\
GPU types & H100 & H100 & H100\\
Training time (hours) & 37 & 95 & 51 \\
Training time with OT (hours) & 38 & 96 & 52 \\
\hline
\textbf{Encoder Details} & & \\
Model Architecture & DDPM++ & DDPM++ & EDM2-S \\
Model Channels & 32 & 32 & 32\\
N$^\circ$ of ResBlocks & 1 & 1 & 2\\
Attention Resolution & [16] & [16] & [16, 8] \\
Channel multiplyer & $[2, 2, 2]$ & $[1, 2, 2, 2]$ & $[1, 2, 3, 4]$\\
$\beta$ regularizer & 10 & 10 & 100 (VE), 90 (LI)\\
Encoder Params & 1.6M & 1.6M & 6M \\
Training time with Encoder (hours) & 49 & 110 & 58\\
\end{tabular}%
}
\caption{Model Configurations and Training Details for ECM on CIFAR10 , FFHQ $64\times46$ and ImageNet $64 \times 64$}
\label{app:tab_ecm}
\end{table}

\subsection{Tuning $\beta$}
\label{app:beta}
In our experiments, we tuned $\beta$ with a coarse grid search with different values with a gap of 10. For iCT on CIFAR10, we initially tested the values $\beta=[10,20,30,40]$, of which $\beta=30$ gave the best performance, then tested also for $\beta=[25,35]$ which did not improve the performance. Similarly, for ECM on CIFAR10, we tested for 
$\beta=[10,20,30,40]$, and after achieving the best performance with $\beta=10$, we tested $\beta=[5,15]$ which did not improve the performance. The tuning was done with the VE kernel and the best values were used also for the LI kernel. We used the best values of $\beta$ also in FashinMNIST and FFHQ without additional tuning. For ImageNet we observed on early runs that $\beta$ needed to be much larger, so we initially tuned for $\beta=[30, 60, 90, 120]$. After achieving the best results with the VE kernel for $\beta=90$, we further tuned for $\beta=[70,80,90,100,110]$ for both VE and LI kernels, and found $\beta=100$ to be the best for VE and $\beta=90$ 
for LI. 

\subsection{The effect of gradient clipping}
\label{app:gcpil}
As our method uses gradient clipping for stability, we test the effect of gradient clipping on the baselines for a fair comparison. We report the results in tables \ref{tab:gclipict} and \ref{tab:gclipecm} for iCT and ECM respectively. While the results generally improve slightly in iCT based methods, it is cleare that the main performance gain in our method is given by the improved coupling.

\begin{table}[h]
\centering
\begin{minipage}{0.48\textwidth}
\centering
\begin{tabular}{lcc}
\hline
Method & 1-step & 2-step \\
\hline
iCT-VE (w/o gc)        & 3.61 & 2.79 \\
iCT-VE (w/ gc)         & 3.49 & 2.57 \\
iCT-LI (w/o gc)        & 3.81 & 2.87 \\
iCT-LI (w/ gc)         & 3.54 & 2.66 \\
iCT-VE-OT (w/o gc)     & 3.28 & 2.66 \\
iCT-VE-OT (w/ gc)      & 3.21 & 2.56 \\
iCT-LI-OT (w/o gc)     & 3.42 & 2.77 \\
iCT-LI-OT (w/ gc)      & 3.18 & 2.63 \\
\rowcolor{gray!15}
iCT-VE-VC (w/ gc) (VCT)     & 2.86 & 2.32 \\
\rowcolor{gray!15}
iCT-VE-LI (w/ gc) (VCT)     & 2.94 & 2.32 \\
\hline
\end{tabular}
\caption{FID performance with/without gc on CIFAR10 for iCT based methods.}
\label{tab:gclipict}
\end{minipage}
\hfill
\begin{minipage}{0.48\textwidth}
\centering
\begin{tabular}{lcc}
\hline
Method & 1-step & 2-step \\
\hline
ECM-VE (w/o gc)        & 3.68 & 2.14 \\
iCT-VE (w/ gc)         & 3.70 & 2.12 \\
iCT-LI (w/o gc)        & 3.65 & 2.14 \\
iCT-LI (w/ gc)         & 3.76 & 2.17 \\
iCT-VE-OT (w/o gc)     & 3.46 & 2.13 \\
iCT-VE-OT (w/ gc)      & 3.45 & 2.12 \\
iCT-LI-OT (w/o gc)     & 3.49 & 2.13 \\
iCT-LI-OT (w/ gc)      & 3.50 & 2.12 \\
\rowcolor{gray!15}
iCT-VE-VC (w/ gc) (VCT)     & 3.26 & 2.02 \\
\rowcolor{gray!15}
iCT-VE-LI (w/ gc) (VCT)    & 2.39 & 2.09 \\
\hline
\end{tabular}
\caption{FID performance with/without gc on CIFAR10 for ECM based methods.}
\label{tab:gclipecm}
\end{minipage}
\end{table}

\subsection{Imagenet training with $100k$ iterations}
\label{app:imgnet100k}
We report here the ImageNet experiments with 100k iterations. For ECM-LI-VC we increased $\beta$ to $\beta = 100$ as $\beta = 90$ diverged during training. Note that for runs with $100k$ iterations, we sometimes encountered divergences for our models with small $\beta$ and sometimes also for the baselines, while it seems to be solved when training for 200k iterations. The 1-step/2-step FID results are reported in Table \ref{tab:imgnet100k}. For the settings with 100k iterations, our method performs similarly or slightly worse than the baseline. We believe this is due to the fact that the encoder for our model requires more iterations to learn the coupling, as demonstrated by the improved results for 200k iterations.
\begin{table}[ht!]
\centering
\begin{tabular}{lcc}
\toprule
\textbf{Method} & \textbf{1-Step FID} & \textbf{2-Step FID} \\
\midrule
ECM-VE           & 5.66 & 3.78 \\
ECM-LI           & 5.63 & 3.48 \\
ECM-VE-OT           & 6.74 & 4.71 \\
ECM-LI-OT           & 6.65 & 4.64 \\
\rowcolor{gray!15}
ECM-VE-VC (VCT)  & 5.67 & 3.67 \\
\rowcolor{gray!15}
ECM-LI-VC (VCT) & 6.34 & 3.77 \\
\bottomrule
\end{tabular}
\caption{Results of our models trained on class conditional IageNet $64\times64$ for $100k$ iterations with batch size 128.}
\label{tab:imgnet100k}
\end{table}

\clearpage`

\section{Comprehensive comparison}
\label{app:fids}
In this section we report a comprehensive comparison with other methods on CIFAR10 and ImageNet $64 \times 64$, in tables \ref{tab:appcifar} and \ref{tab:fid_vs_nfe}.
\begin{table}[ht!]
\centering
\caption{FID, NFE and \# param.\ on CIFAR-10. Bold indicates the best result for each category and NFE.}
\begin{tabular}{lccc}
\toprule
Method & NFE & FID & \# param.\ (M) \\
\midrule
\multicolumn{4}{l}{\textbf{Diffusion models}} \\
EDM \citep{karras2022elucidating}                & 35   & 1.97           & 55.7  \\
PFGM++ \citep{xu2023pfgm++}               & 35   & \textbf{1.91}  & 55.7  \\
DDPM \citep{ho2020denoising}                  & 1000 & 3.17           & 35.7  \\
LSGM \citep{vahdat2021score}              & 147  & 2.10           & 475   \\
\midrule
\multicolumn{4}{l}{\textbf{Consistency models}} \\
\multicolumn{4}{l}{\quad\textit{1‑step}} \\
iCT \citep{songimproved}         & 1    & 2.83           & 56.4  \\
iCT‑deep \citep{songimproved}    & 1    & 2.51           & 112   \\
CTM \citep{kimconsistency} (w/o GAN)         & 1    & 5.19           & 55.7  \\
ECM \citep{geng2024consistency}                 & 1    & 3.60           & 55.7  \\
TCM \citep{lee2024truncated}                           & 1    & \textbf{2.46}  & 55.7  \\
sCT \cite{lu2024simplifying} & 1 & 2.85 & - \\
\rowcolor{gray!15}
iCT-VE-VC (VCT) & 1 & 2.86 & 57.3 \\
\rowcolor{gray!15}
iCT-LI-VC (VCT) & 1 & 2.94 & 57.3 \\
\rowcolor{gray!15}
ECM-VE-VC (VCT) & 1 & 3.26 & 57.3 \\
\rowcolor{gray!15}
ECM-LI-VC (VCT) & 1 & 3.39 & 57.3 \\
\multicolumn{4}{l}{\quad\textit{2‑step}} \\
iCT \citep{songimproved}         & 2    & 2.46           & 56.4  \\
iCT‑deep \citep{songimproved}    & 2    & 2.24           & 112   \\
ECM \citep{geng2024consistency}                 & 2    & 2.11           & 55.7  \\
TCM \citep{lee2024truncated}                           & 2    & 2.05  & 55.7  \\
sCT \cite{lu2024simplifying} & 2 & 2.06 & - \\
\rowcolor{gray!15}
iCT-VE-VC (VCT) & 2 & 2.32 & 57.3 \\
\rowcolor{gray!15}
iCT-LI-VC (VCT) & 2 & 2.32 & 57.3 \\
\rowcolor{gray!15}
ECM-VE-VC (VCT) & 2 & \textbf{2.02} & 57.3 \\
\rowcolor{gray!15}
ECM-LI-VC (VCT) & 2 & 2.09 & 57.3 \\
\midrule
\multicolumn{4}{l}{\textbf{Variational score distillation}} \\
DMD \citep{yin2024one}                 & 1    & 3.77           & 55.7  \\
Diff‑Instruct \citep{luo2023diff}        & 1    & 4.53           & 55.7  \\
SiD \citep{zhou2024score}                 & 1    & \textbf{1.92}  & 55.7  \\
\midrule
\multicolumn{4}{l}{\textbf{Knowledge distillation}} \\
KD \citep{luhman2021knowledge}                & 1    & 9.36           & 35.7  \\
DSNO \citep{zheng2023fast}              & 1    & 3.78           & 65.8  \\
TRACT \citep{berthelot2023tract}          & 1    & 3.78           & 55.7  \\
                                     & 2    & 3.32  & 55.7  \\
PD \citep{salimansprogressive}            & 1    & 9.12           & 60.0  \\
                                     & 2    & 4.51           & 60.0  \\
CD (LPIPS) \citep{song2023consistency} & 1 & 3.55 & - \\
                                       & 2 & 2.93 & - \\
sCD \citep{lu2024simplifying} & 1 & 3.66 & - \\
                              & 2 & \textbf{2.52} & - \\

\bottomrule
\end{tabular}
\label{tab:appcifar}
\end{table}

\begin{table}[ht]
\centering
\caption{FID, NFE and $\#$ param. on ImageNet $64\times64$. Dotted
lines separate results by $\#$ param. Bold indicates the best result for each category and NFE. Our models, marked with a $^*$, use batch size 128 while equivalent models based on ECM use batch size 1024.}
\label{tab:fid_vs_nfe}
\begin{tabular}{l c c r}
\toprule
Method & NFE & FID & \# param.\ (M) \\
\midrule
\multicolumn{4}{l}{\textbf{Diffusion models}} \\
\quad EDM2‑S \citep{karras2024analyzing} & 63 & 1.58 &  280 \\
\quad EDM2‑XL \citep{karras2024analyzing} & 63 & 1.33 & 1119 \\
\midrule
\multicolumn{4}{l}{\textbf{Consistency models}} \\
\cmidrule(l){1-4}
\multicolumn{4}{l}{\quad\textit{1‑step}} \\
\quad iCT \citep{songimproved}               & 1 & 4.02 &  296 \\
\quad iCT‑deep \citep{songimproved}          & 1 & 3.25 &  592 \\
\quad ECM \citep{geng2024consistency} (EDM2‑S)            & 1 & 4.05 &  280 \\
\quad TCM \citep{lee2024truncated}(EDM2‑S)                         & 1 & \textbf{2.88} &  280 \\
\quad sCT \citep{lu2024simplifying} (EDM2-S) & 1 & 3.23 & 280 \\
\quad ECM \citep{geng2024consistency} (EDM2‑XL)           & 1 & 2.49 & 1119 \\
\quad TCM \citep{lee2024truncated} (EDM2‑XL)                        & 1 & 2.20 & 1119 \\
\quad sCT \citep{lu2024simplifying} (EDM2-XL) & 1 & \textbf{2.04} & 1119 \\
\rowcolor{gray!15}
\quad ECM-VE-VC (Ours, EDM2-S)$^*$  & 1 & 5.08 & 286\\
\rowcolor{gray!15}
\quad ECM-LI-VC (Ours, EDM2-S)$^*$  & 1 & 4.93 & 286\\
\cmidrule(l){1-4}
\multicolumn{4}{l}{\quad\textit{2‑step}} \\
\quad iCT \citep{songimproved}              & 2 & 3.20 &  296 \\
\quad iCT‑deep \citep{songimproved}        & 2 & 2.77 &  592 \\
\quad ECM \citep{geng2024consistency} (EDM2‑S)            & 2 & 2.79 &  280 \\
\quad TCM \citep{lee2024truncated}(EDM2‑S)                         & 2 & \textbf{2.31} &  280 \\
\quad sCT \citep{lu2024simplifying} (EDM2-S) & 2 & 2.93 & 280 \\
\quad ECM \citep{geng2024consistency}(EDM2‑XL)           & 2 & 1.67 & 1119 \\
\quad TCM \citep{lee2024truncated}(EDM2‑XL)                        & 2 & 1.62 & 1119 \\
\quad sCT \citep{lu2024simplifying} (EDM2-XL) & 2 & \textbf{1.48} & 1119 \\
\rowcolor{gray!15}
\quad ECM-VE-VC (Ours, EDM2-S)$^*$ & 2 & 3.15 & 286\\
\rowcolor{gray!15}
\quad ECM-LI-VC (Ours, EDM2-S)$^*$ & 2 & 3.07 & 286\\
\midrule
\multicolumn{4}{l}{\textbf{Variational score distillation}} \\
\quad DMD2 w/o GAN \citep{yin2024one}           & 1 & 2.60 &  296 \\
\quad Diff‑Instruct \citep{luo2023diff}           & 1 & 5.57 &  296 \\
\quad EMD‑16 \citep{xie2024distillation}              & 1 & 2.20 &  296 \\
\quad Moment Matching \citep{salimans2024multistep}    & 1 & 3.00 &  400 \\
\quad                                            & 2 & 3.86 &  400 \\
\quad SiD \citep{zhou2024score}                   & 1 & \textbf{1.52} &  296 \\
\midrule
\multicolumn{4}{l}{\textbf{Knowledge distillation}} \\
\quad DSNO \citep{zheng2023fast}                & 1 & 7.83 &  329 \\
\quad TRACT \citep{berthelot2023tract}              & 1 & 7.43 &  296 \\
\quad                                            & 2 & 4.97 &  296 \\
\quad PD \citep{salimansprogressive}                   & 1 & 15.4 &  296 \\
\quad                                            & 2 & 8.95 &  296 \\
\quad CD (LPIPS) \citep{song2023consistency} & 1 & 6.20 & - \\
                                       & 2 & 4.70 & - \\
\quad sCD \citep{lu2024simplifying} (EDM2-XL) & 1 & \textbf{2.44} & 1119 \\
                              & 2 & \textbf{1.66} & 1119 \\
\bottomrule
\end{tabular}
\label{tab:fidimnet}
\end{table}

\clearpage

\section{Toy experiments}
\label{app:toy}
\begin{figure}[ht!]
    \centering
    \includegraphics[width=0.7\linewidth]{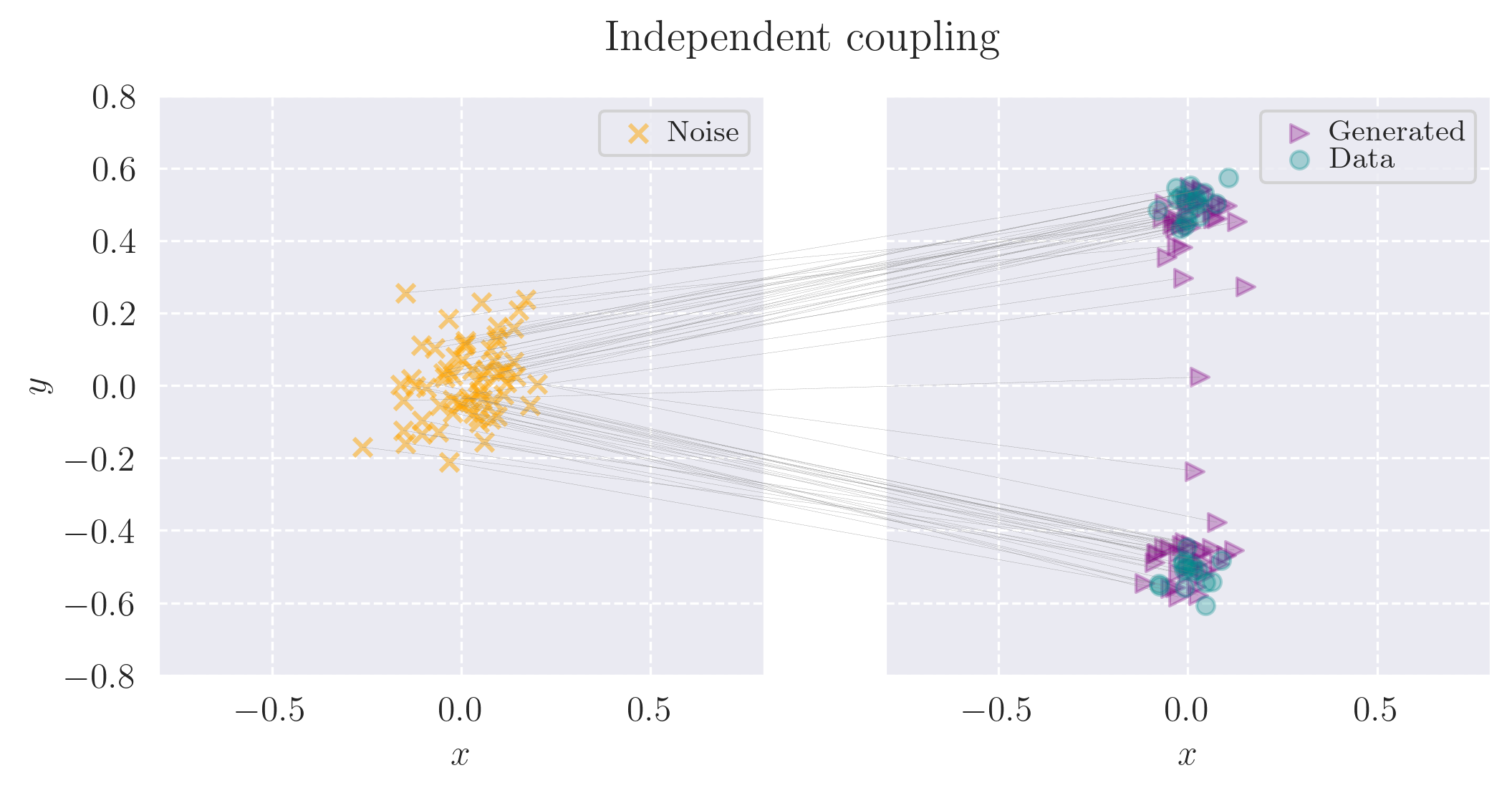}
    \includegraphics[width=0.7\linewidth]{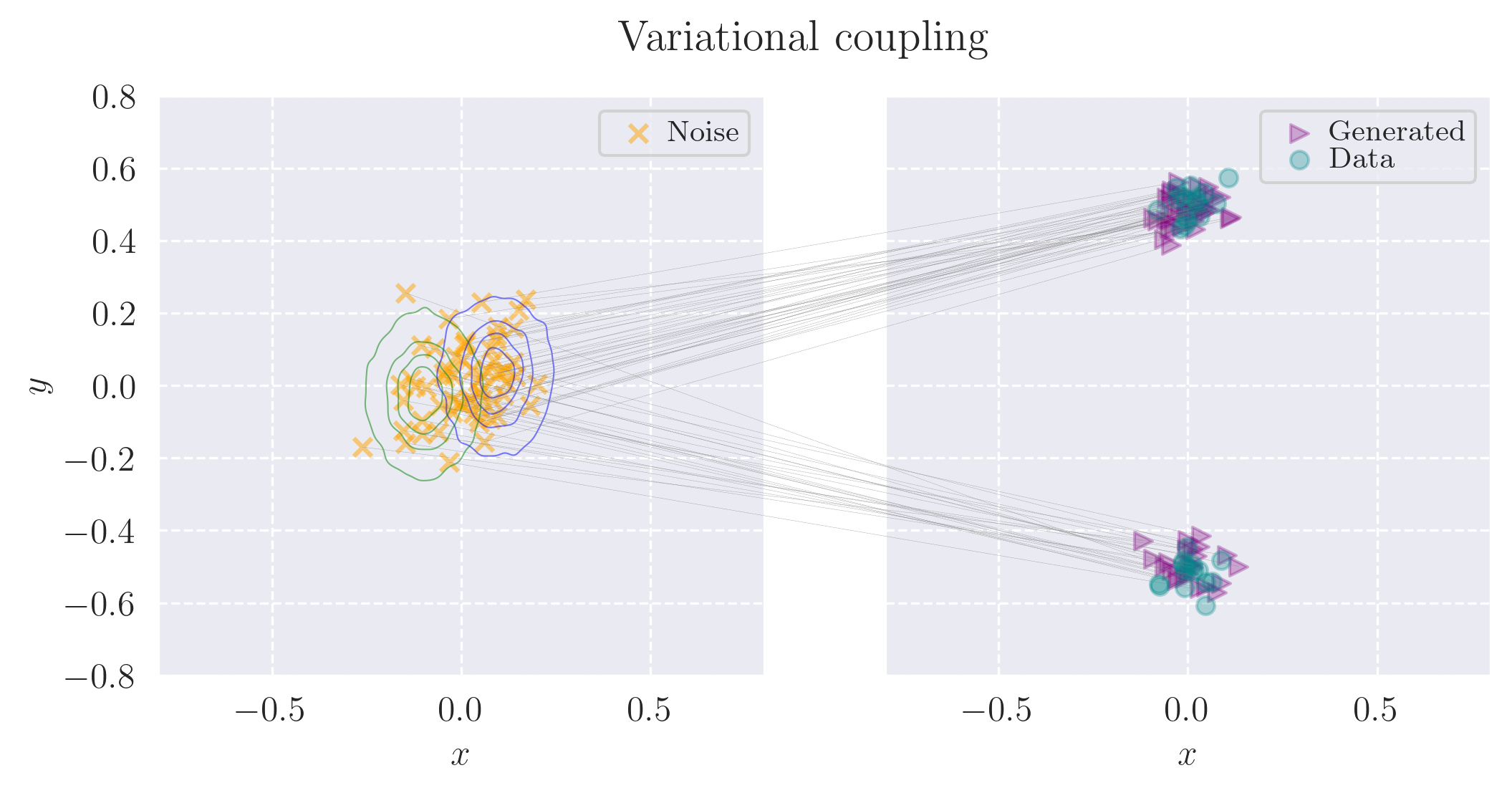}
    \caption{2-step generation result on the toy data, with $t=0.07$.}
    \label{app:toy_samples_2}
\end{figure}

To gain a visual understanding of the benefits of the variational coupling, we use the model to learn the distribution of a mixture of two Gaussians, with means $\mu_1=(0, 0.5)$ and $\mu_2=(0, -0.5)$, and standard deviation $\sigma=0.05$. We use iCT-LI so that the perturbed data reaches the prior even with small $\sigma_{\mathrm{max}}$, with $\sigma_\mathrm{min}=0.002$, $\sigma_{\mathrm{max}}=0.1$ and $\sigma_{\mathrm{data}}=0.05$. The models are trained for 40k iterations, with $s_0=10$ and $s_1=80$, and EMA rate 0.999. We use a simple four-layers MLP with GeLU activation and Positional time embedding, with batch size 256 and learning rate $1e^{-4}$. For iCT-LI-VC we use $\beta=0.001$. The results are shown in figures \ref{fig:toy_samples_1} and \ref{app:toy_samples_2} (one and two step generation respectively).

\clearpage
\section{Model diagram}
\label{app:diag}
\begin{figure}[ht!]
    \centering
    \includegraphics[width=\linewidth]{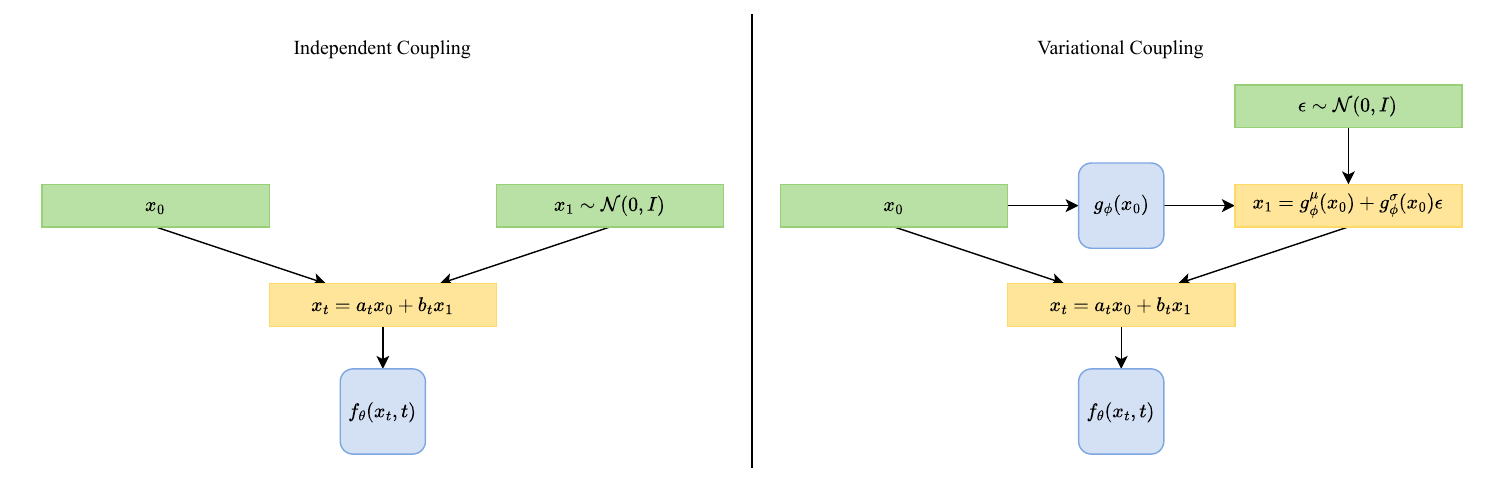}
    \caption{Diagram for Consistency Training with independent coupling (left) and variational coupling (right).}
    \label{app:graph}
\end{figure}

In figure \ref{app:graph} we show the difference between the forward process for standard Consistency Training and for our method with learned noise coupling, for a given time step $t$ and transition kernel characterized by the coefficients $a_t$ and $b_t$.
\section{Qualitative Results}
\label{app:qualitative}
\begin{figure}[ht!]
    \centering
    \includegraphics[width=0.5\linewidth]{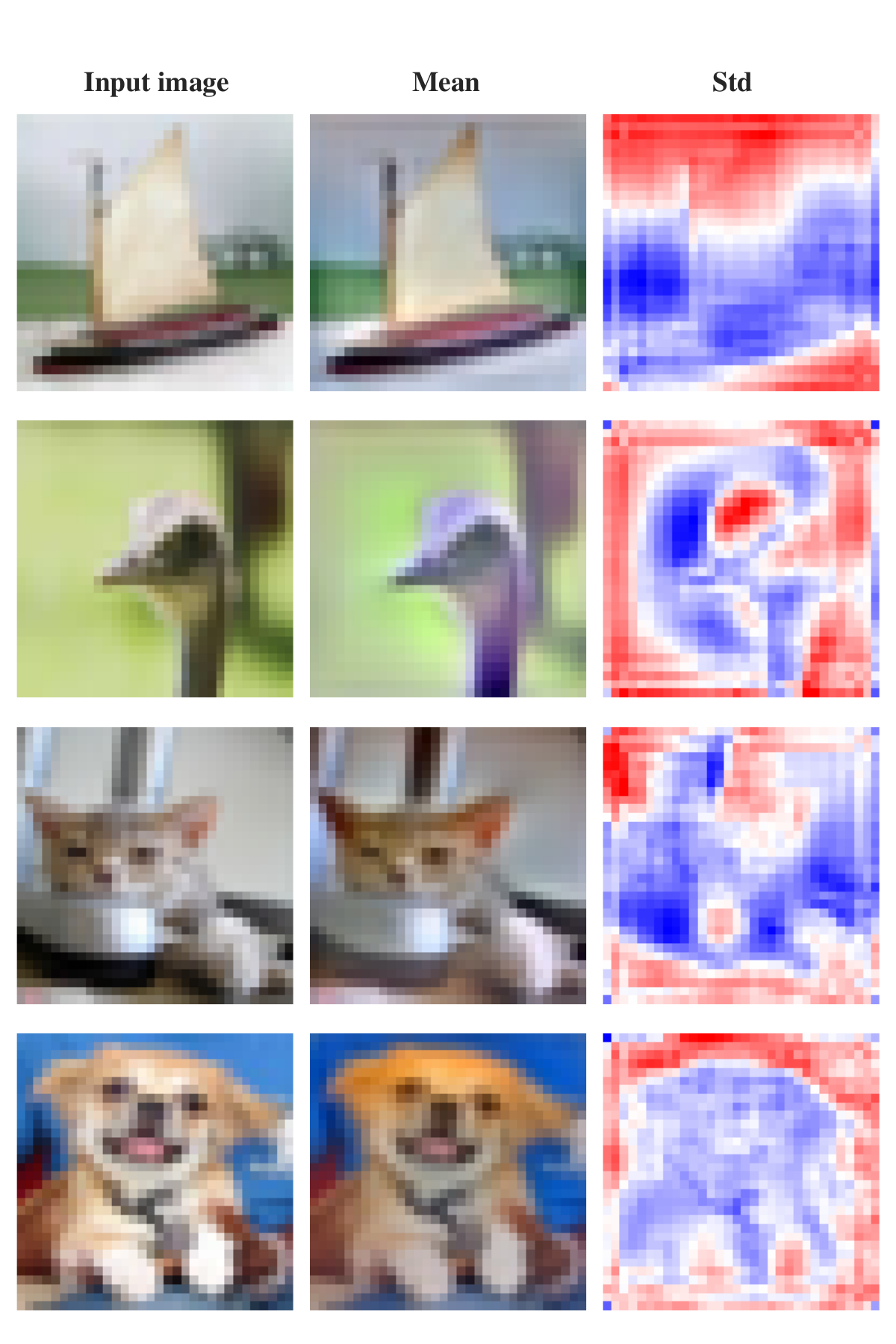}
    \caption{Visualization of the predicted mean and standard deviation for a trained iCT-VE-VC model for different input images. For visualization purpose, we perform min-max rescaling for the predicted mean and standard deviation, as they tend to have most values close to zero and one respectively. We also turn the predicted 3 channels standard deviations to a single channel with grayscale transform.}
    \label{fig:rec}
\end{figure}

\begin{figure}[ht!]
    \centering
    \includegraphics[width=0.49\linewidth]{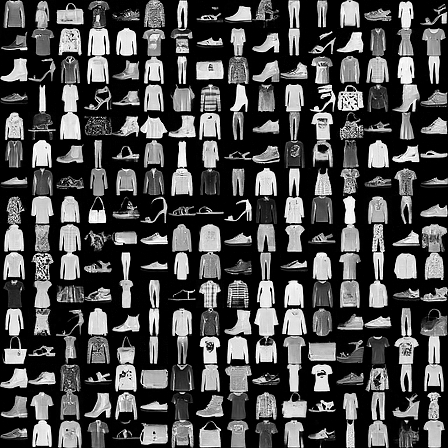}
    \includegraphics[width=0.49\linewidth]{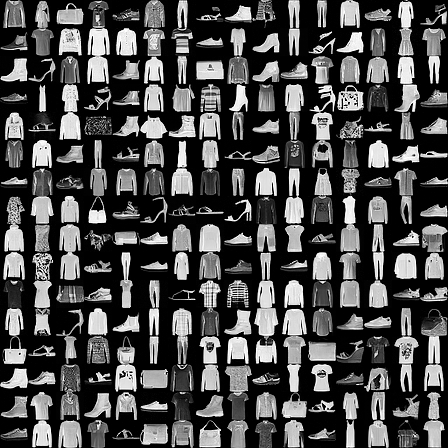}
    \caption{1-step (FID=$3.62$, left) and 2-step (FID=$2.22$, right) generation from iCT-LI-VC trained on FashionMNIST.}
    \label{app:samples_fmnist}
\end{figure}

\begin{figure}[ht!]
    \centering
    \includegraphics[width=0.49\linewidth]{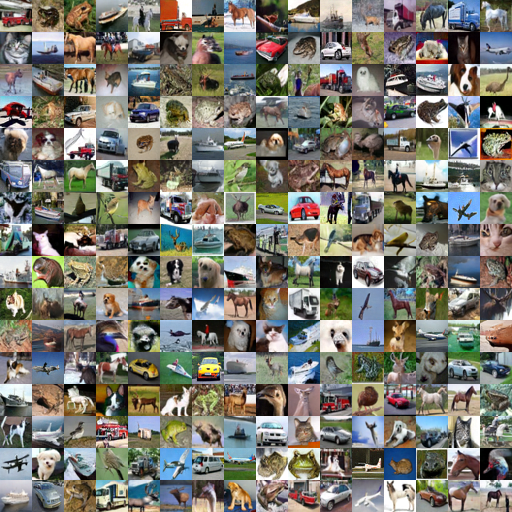}
    \includegraphics[width=0.49\linewidth]{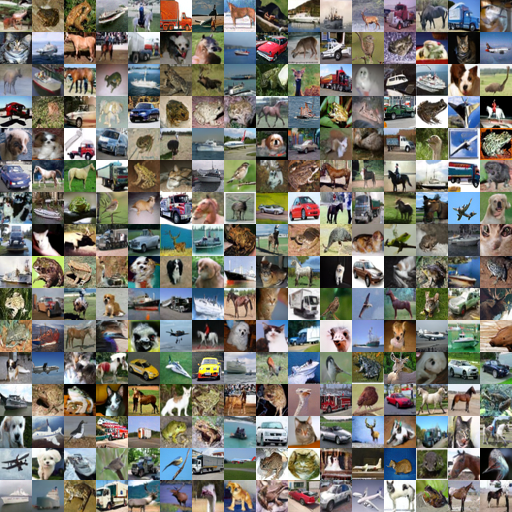}
    \caption{1-step (FID=$2.86$, left) and 2-step (FID=$2.32$, right) generation from iCT-VE-VC trained on CIFAR10.}
    \label{app:samples_cifar1}
\end{figure}

\begin{figure}[ht!]
    \centering
    \includegraphics[width=0.49\linewidth]{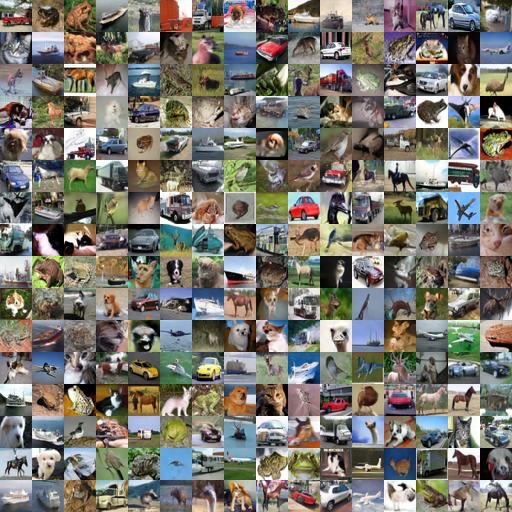}
    \includegraphics[width=0.49\linewidth]{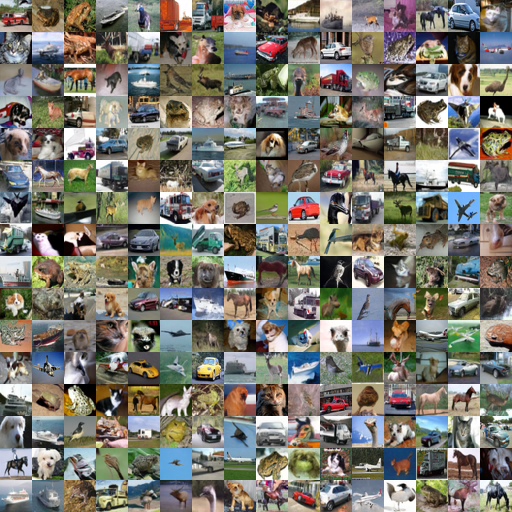}
    \caption{1-step (FID=$3.26$, left) and 2-step (FID=$2.02$, right) generation from ECM-VE-VC trained on CIFAR10.}
    \label{app:samples_cifar2}
\end{figure}

\begin{figure}[ht!]
    \centering
    \includegraphics[width=0.49\linewidth]{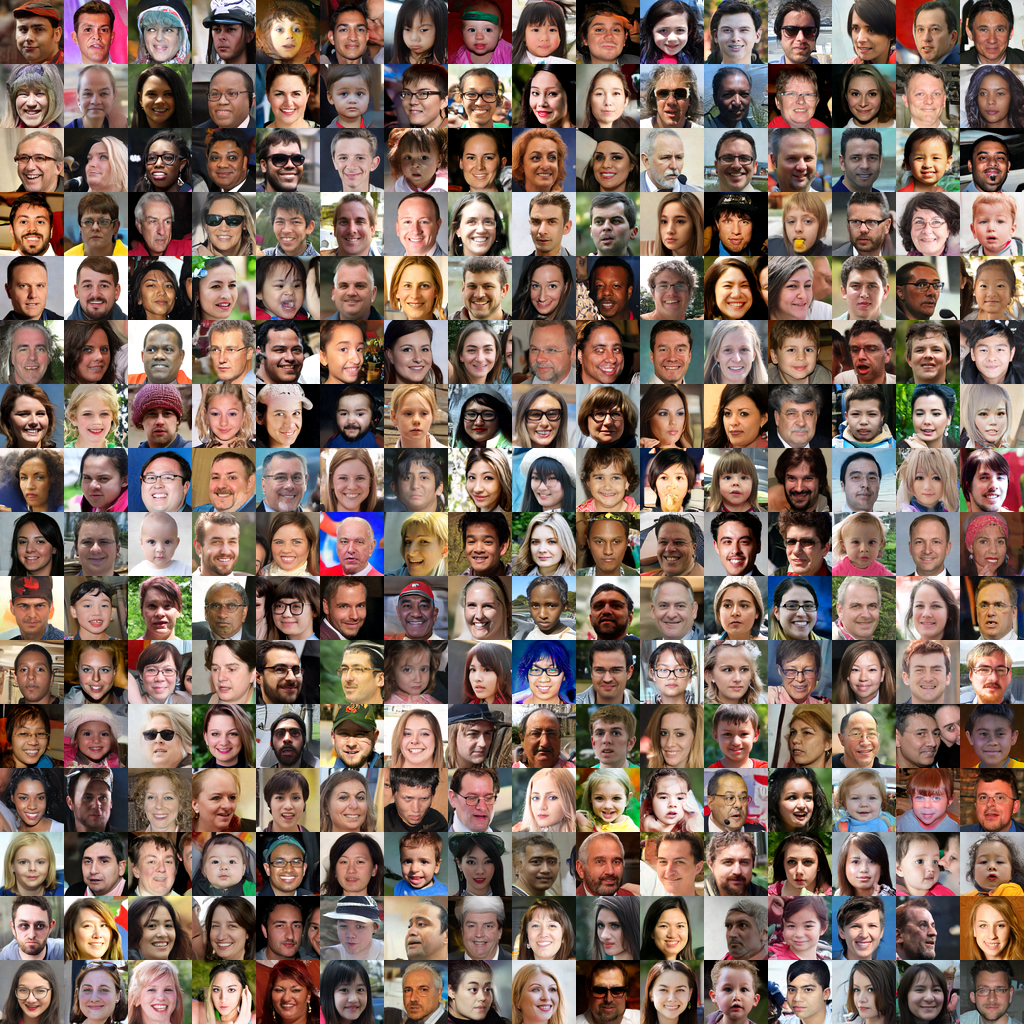}
    \includegraphics[width=0.49\linewidth]{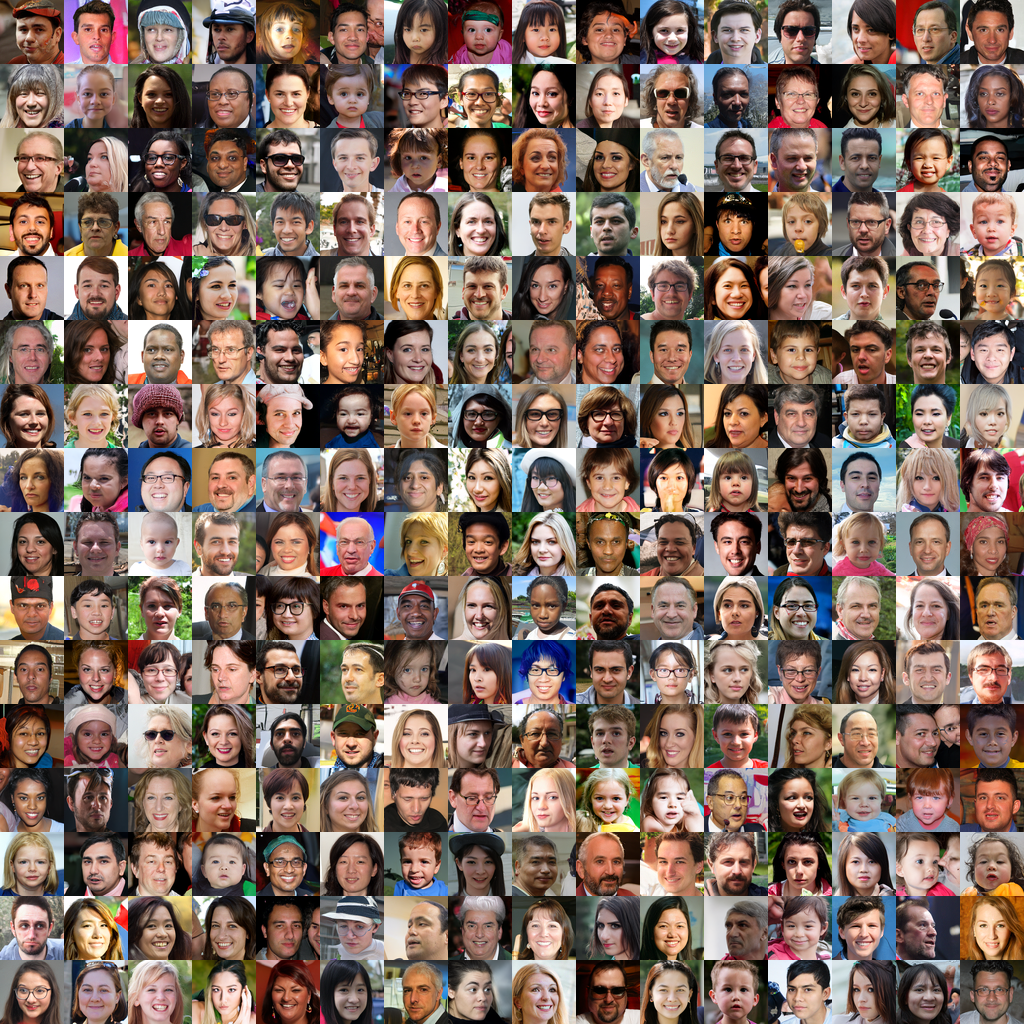}
    \caption{1-step (FID=$5.47$, left) and 2-step (FID=$4.16$, right) generation from ECM-VE-VC trained on FFHQ$64\times64$.}
    \label{app:samples_ffhq}
\end{figure}

\begin{figure}[ht!]
    \centering
    \includegraphics[width=0.49\linewidth]{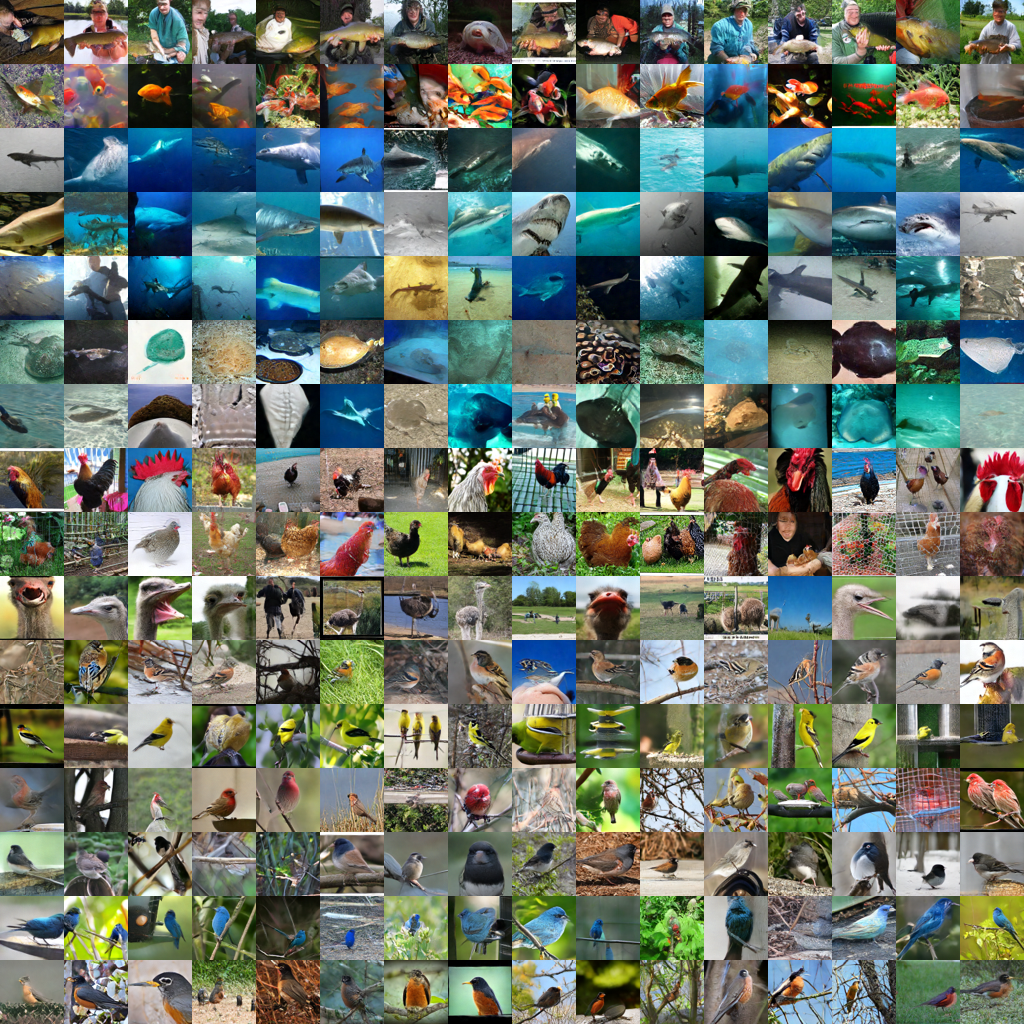}
    \includegraphics[width=0.49\linewidth]{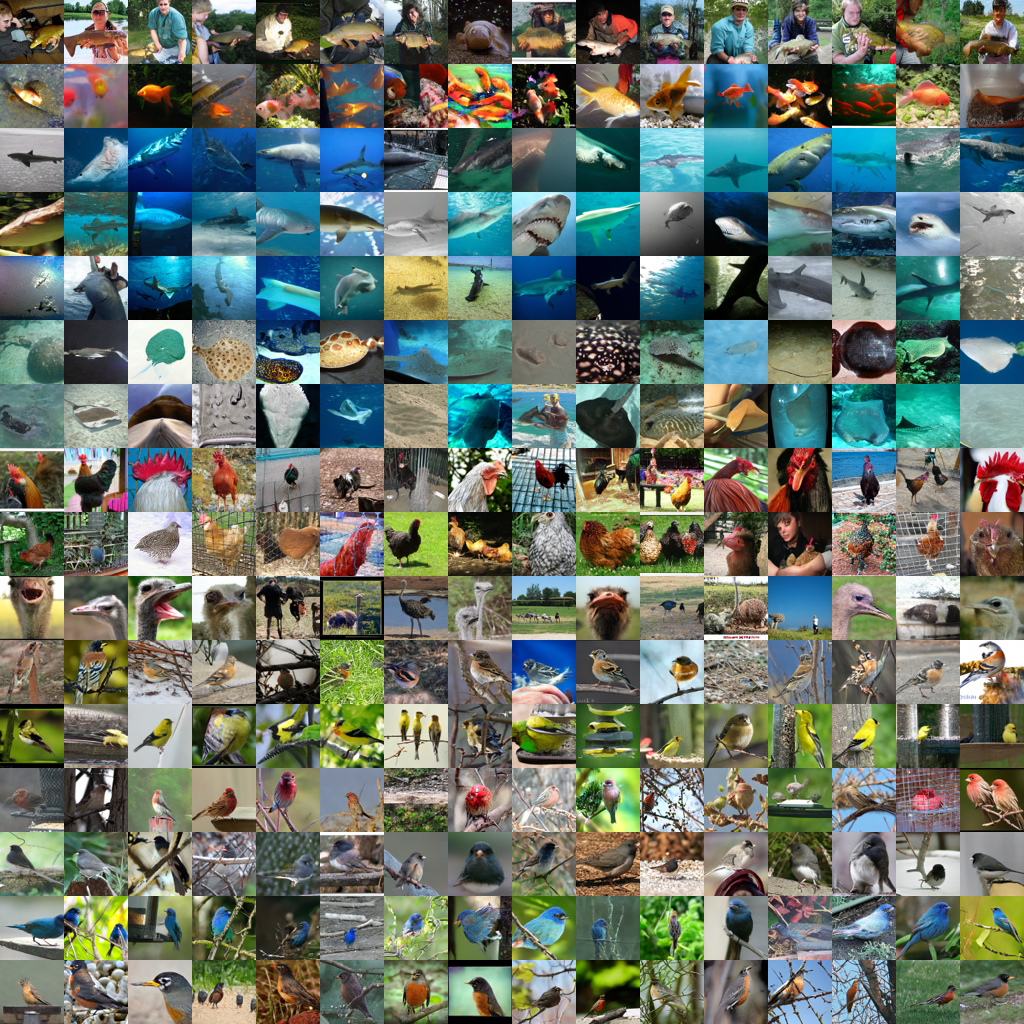}
    \caption{1-step (FID=$4.93$, left) and 2-step (FID=$3.07$, right) generation from ECM-LI-VC trained on class conditional ImageNet $64\times64$.}
    \label{app:samples_imnet}
\end{figure}

Here we report samples from our best models, iCT-LI-VC for FashionMNIST (figure \ref{app:samples_fmnist}), iCT-VE-VC and ECM-VE-VC on CIFAR10 (figures \ref{app:samples_cifar1} and \ref{app:samples_cifar2}), ECM-VE-VC on FFHQ $64\times 64$ (figure \ref{app:samples_ffhq}) and ECM-LI-VC on class conditional Imagenet $64\times64$ (figure \ref{app:samples_imnet}). In figure \ref{fig:rec}, we show the mean and standard deviation learned by the encoder for some images from the CIFAR10 dataset. While the values are very close to a standard Gaussian, the model still retains information from the original input. 

\end{document}

%% file: math_commands.tex
\usepackage{amsmath,amsfonts,bm}

\def\eqref#1{(\ref{#1})}

\def\1{\bm{1}}

\def\rvx{{\mathbf{x}}}

\def\rmI{{\mathbf{I}}}

\def\rmzr{{\mathbf{0}}}

\def\vx{{\bm{x}}}

\DeclareMathAlphabet{\mathsfit}{\encodingdefault}{\sfdefault}{m}{sl}
\SetMathAlphabet{\mathsfit}{bold}{\encodingdefault}{\sfdefault}{bx}{n}

\newcommand{\KL}{\mathcal{D}_{\mathrm{KL}}}

